# The Mechanics of $n$-Player Differentiable Games

**David Balduzzi** [1] **Sébastien Racanière** [1] **James Martens** [1] **Jakob Foerster** [2] **Karl Tuyls** [1] **Thore Graepel** [1]

## Abstract

The cornerstone underpinning deep learning is the guarantee that gradient descent on an objective converges to local minima. Unfortunately, this guarantee fails in settings, such as generative adversarial nets, where there are multiple interacting losses. The behavior of gradient-based methods in games is not well understood – and is becoming increasingly important as adversarial and multi-objective architectures proliferate. In this paper, we develop new techniques to understand and control the dynamics in general games. The key result is to decompose the second-order dynamics into two components. The first is related to potential games, which reduce to gradient descent on an implicit function; the second relates to *Hamiltonian games*, a new class of games that obey a conservation law, akin to conservation laws in classical mechanical systems. The decomposition motivates *Symplectic Gradient Adjustment* (SGA), a new algorithm for finding stable fixed points in general games. Basic experiments show SGA is competitive with recently proposed algorithms for finding stable fixed points in GANs – whilst at the same time being applicable to – and having guarantees in – much more general games.

## 1. Introduction

Recent progress in machine learning is heavily dependent on using gradient descent, applied to optimize the parameters of models with respect to a (single) objectives. A basic result is that gradient descent converges to a local minimum of the objective under a broad range of conditions (Lee et al., 2017). However, there is a rapidly growing set of powerful models that do not optimize a single objective, including: generative adversarial networks (Goodfellow et al., 2014), proximal gradient TD learning (Liu et al., 2016), multi-level optimization (Pfau & Vinyals, 2016), synthetic gradients (Jaderberg et al., 2017), hierarchical reinforcement learning (Wayne & Abbott, 2014; Vezhnevets et al., 2017), curiosity (Pathak et al., 2017), and imaginative agents (Racanière et al., 2017). In effect, the models are trained via games played by cooperating and competing modules.

No-regret algorithms such as gradient descent are guaranteed to converge to coarse correlated equilibria in games (Stoltz & Lugosi, 2007). However, the dynamics do not converge to Nash equilibria – and do not even stabilize – in general (Mertikopoulos et al., 2018). Concretely, cyclic behaviors emerge even in simple cases, see example 1.

This paper presents an analysis of the second-order structure of game dynamics that allows to identify two classes of games, potential and Hamiltonian, that are easy to solve separately. We then derive *symplectic gradient adjustment* (SGA), a method for finding stable fixed points in games. SGA's performance is evaluated in basic experiments.

**Background and problem description.** Procedures that converge to Nash equilibria have been found for restricted game classes: potential games, 2-player zero-sum games and a few others (Hu & Wellman, 2003; Hart & Mas-Colell, 2013). Finding equilibria can be reformulated as a nonlinear complementarity problem, but these are 'hopelessly impractical to solve' in general (Shoham & Leyton-Brown, 2008) because the problem is PPAD hard (Daskalakis et al., 2009).

Players are primarily neural nets in our setting. We therefore restrict to gradient-based methods (game-theorists have considered a much broader range of techniques). Losses are not necessarily convex in *any* of their parameters, so Nash equilibria do not necessarily exist. Leaving existence aside, finding Nash equilibria is analogous to, but much harder than, finding global minima in neural nets – which is not realistic with gradient-based methods.

There are (at least) three problems with gradient descent in games. Firstly, the potential existence of cycles (recurrent dynamics) implies there are no convergence guarantees, see example 1 and Mertikopoulos et al. (2018). Secondly, even when gradient descent converges, the rate may be too slow in practice because 'rotational forces' necessitate extremely small learning rates (see figure 3). Finally, since there is no single objective, there is no way to measure progress. Application-specific proxies have been proposed, for example the inception score for GANs (Salimans et al., 2016), but these are little help during training – the inception score is no substitute for looking at samples.

**Outline and summary of main contributions.** We start with the well-known case of a zero-sum bimatrix game: example 1. It turns out that the dynamics (that is, the dynamics under simultaneous gradient descent) can be reformulated via Hamilton's equations. The cyclic behavior arises because the dynamics live on the level sets of the Hamiltonian. More directly useful, gradient descent on the Hamiltonian converges to a Nash equilibrium.

Lemma 1 shows that the Hessian of any game decomposes into symmetric and antisymmetric components. There are thus two 'pure' cases: when only the symmetric component is present, or only the antisymmetric. The first case, known as potential games (Monderer & Shapley, 1996), have been intensively studied because they are exactly the games where gradient descent *does* converge.

The second case, Hamiltonian[1] games, were not studied previously, probably because they coincide with zero-sum games in the bimatrix case (or constant-sum, depending on the constraints). Zero-sum and Hamiltonian games differ when the losses are not bilinear or there are more than two players. Hamiltonian games are important because (i) they are easy to solve and (ii) general games combine potential-like and Hamiltonian-like dynamics. The concept of a zero-sum game is too loose to be useful when there are many players: any $n$-player game can be reformulated as a zero-sum $(n+1)$-player game where $\ell_{n+1} = -\sum_{i=1}^{n} \ell_i$. Zero-sum games are as complicated as general-sum games. Theorem 3 shows that Hamiltonian games obey a conservation law – which also provides the key to solving them, by gradient descent on the conserved quantity.

The general case, neither potential nor Hamiltonian, is more difficult and the focus of the remainder of the paper. Section 3 proposes *symplectic gradient adjustment (SGA)*, a gradient-based method for finding stable fixed points in general games. Appendix C contains TensorFlow code to compute the adjustment. The algorithm computes two Hessian-vector products, at a cost of two iterations of backprop. SGA satisfies a few natural desiderata: $D1$ it is compatible with the original dynamics and $D2$, $D3$ it is guaranteed to find stable equilibria in potential and Hamiltonian games.

For general games, correctly picking the *sign* of the adjustment (whether to add or subtract) is critical since it determines the behavior near stable and unstable equilibria. Section 2.3 defines stable equilibria and relates them to local Nash equilibria. Lemma 10 then shows how to set the sign so as to converge to stable fixed points. Correctly aligning

SGA allows higher learning rates and faster, more robust convergence, see theorem 7 and experiments in section 4.

Section 4 investigates a basic GAN setup from Metz et al. (2017), that tests for mode-collapse and mode hopping. Whereas simultaneous gradient descent completely fails; the symplectic adjustment leads to rapid convergence – slightly improved by correctly choosing the sign of the adjustment. Finally, section 3.5 applies the same criterion to align consensus optimization (Mescheder et al., 2017), preventing it from converging to unstable equilibria and (slightly) improving performance, figure 9 in the appendix.

**Caveat.** The behavior of SGA near fixed points that are neither negative nor positive semi-definite is not analysed. On the one hand, it was only recently shown that gradient descent behaves well near saddles when optimizing a single objective (Lee et al., 2016; 2017). On the other hand, Newton's method is attracted to saddles, see analysis and recently proposed remedy in Dauphin et al. (2014). Studying indefinite fixed points is deferred to future work.

**Related work.** Convergence to Nash equilibria in two-player games was studied in Singh et al. (2000). WoLF (Win or Learn Fast) converges to Nash equilibria in two-player two-action games (Bowling & Veloso, 2002). Extensions include weighted policy learning (Abdallah & Lesser, 2008) and GIGA-WoLF (Bowling, 2004). There has been almost no work on convergence to fixed points in general games. Optimistic mirror descent approximately converges in two-player bilinear zero-sum games (Daskalakis et al., 2018), a special case of Hamiltonian games. In more general settings it converges to coarse correlated equilibria.

There has been interesting recent work on convergence in GANs. Heusel et al. (2017) propose a two-time scale methods to find Nash. However, it likely scales badly with the number of players. Nagarajan & Kolter (2017) prove convergence for some algorithms, but under very strong assumptions Mescheder (2018). Consensus optimization (Mescheder et al., 2017) is discussed in section 3. Learning with opponent-learning awareness (LOLA) infinitesimally modifies the objectives of players to take into account their opponents' goals (Foerster et al., 2018). However, there are no guarantees that LOLA converges and in general it may modify the fixed-points of the game.

**Notation.** Dot products are written as $\mathbf{v}^\intercal \mathbf{w}$ or $\langle \mathbf{v}, \mathbf{w} \rangle$. The angle between vectors is $\theta(\mathbf{v}, \mathbf{w})$. Positive definiteness is denoted $\mathbf{S} \succ 0$. Omitted proofs are in appendix B.

## 2. The infinitesimal structure of games

In contrast to the classical formulation of games, we do not constrain the parameter sets (e.g. to the probability simplex) or require losses to be convex in the corresponding players'

---

[1] Lu (1992) defined an unrelated notion of Hamiltonian game.

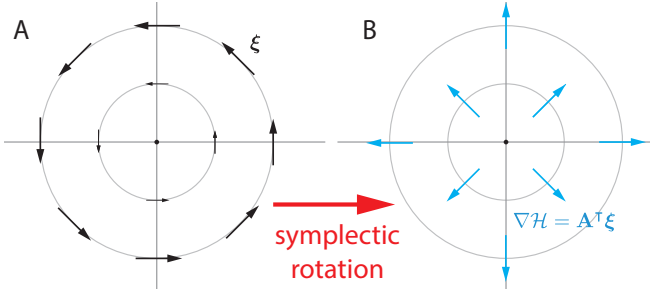

*Figure 1.* **(A)** $\boldsymbol{\xi}$ cycles around the origin. **(B)** gradient descent on $\mathcal{H}$ converges to Nash.

parameters. Players could be interacting neural nets such as GANs (Goodfellow et al., 2014).

**Definition 1.** *A game is a set of players* $[n] = \{1, \dots, n\}$ *and twice continuously differentiable losses* $\{\ell_i : \mathbb{R}^d \to \mathbb{R}\}_{i=1}^n$. *Parameters are* $\mathbf{w} = (\mathbf{w}_1, \dots, \mathbf{w}_n) \in \mathbb{R}^d$ *with* $\mathbf{w}_i \in \mathbb{R}^{d_i}$ *where* $\sum_{i=1}^n d_i = d$. . *Player i controls* $\mathbf{w}_i$.

It is sometimes convenient to write $\mathbf{w} = (\mathbf{w}_i, \mathbf{w}_{-i})$ where $\mathbf{w}_{-i}$ concatenates the parameters of all the players other than the $i^{\text{th}}$, which is placed out of order by abuse of notation.

The *simultaneous gradient* is the gradient of the losses with respect to the parameters of the respective players:

$$\boldsymbol{\xi}(\mathbf{w}) = (\nabla_{\mathbf{w}_1} \ell_1, \dots, \nabla_{\mathbf{w}_n} \ell_n) \in \mathbb{R}^d.$$

By the **dynamics** of the game, we mean following the negative of the vector field $\boldsymbol{\xi}$ with infinitesimal steps. There is no reason to expect $\boldsymbol{\xi}$ to be the gradient of a *single* function in general, and therefore no reason to expect the dynamics to converge to a fixed point.

### 2.1. Warmup: Hamiltonian mechanics in games

The next example illustrates the essential problem with gradients in games and the key insight motivating our approach.

**Example 1** (Conservation of energy in a zero-sum unconstrained bimatrix game)**.** *Zero-sum games, where* $\sum_{i=1}^n \ell_i \equiv 0$, *are well-studied. The zero-sum game*

$$\ell_1(\mathbf{x}, \mathbf{y}) = \mathbf{x}^\intercal A \mathbf{y} \quad \text{and} \quad \ell_2(\mathbf{x}, \mathbf{y}) = -\mathbf{x}^\intercal A \mathbf{y}$$

*has Nash equilibrium at* $(\mathbf{x}, \mathbf{y}) = (\mathbf{0}, \mathbf{0})$. *The simultaneous gradient* $\boldsymbol{\xi}(\mathbf{x}, \mathbf{y}) = (A\mathbf{y}, -A^\intercal \mathbf{x})$ *rotates around the Nash, see figure 1.*

*The matrix A admits singular value decomposition (SVD)* $A = \mathbf{U}^\intercal \mathbf{D} \mathbf{V}$. *Changing to coordinates* $\mathbf{u} = \mathbf{D}^{\frac{1}{2}} \mathbf{U} \mathbf{x}$ *and* $\mathbf{v} = \mathbf{D}^{\frac{1}{2}} \mathbf{V} \mathbf{y}$ *gives* $\ell_{1/2}(\mathbf{u}, \mathbf{v}) = \pm \mathbf{u}^\intercal \mathbf{v}$. *Introduce the Hamiltonian*

$$\mathcal{H} = \frac{1}{2} \left( \|\mathbf{u}\|_2^2 + \|\mathbf{v}\|_2^2 \right) = \frac{1}{2} \left( \mathbf{x}^\intercal \mathbf{U}^\intercal \mathbf{D} \mathbf{U} \mathbf{x} + \mathbf{y}^\intercal \mathbf{V}^\intercal \mathbf{D} \mathbf{V} \mathbf{y} \right).$$

*Remarkably, the dynamics can be reformulated via Hamilton's equations in the coordinates given by the SVD of A:*

$$\boldsymbol{\xi}(\mathbf{u}, \mathbf{v}) = \left( \frac{\partial \mathcal{H}}{\partial \mathbf{v}}, -\frac{\partial \mathcal{H}}{\partial \mathbf{u}} \right).$$

*Vector field* $\boldsymbol{\xi}$ *cycles around the equilibrium because* $\boldsymbol{\xi}$ *conserves the Hamiltonian's level sets (i.e.* $\langle \boldsymbol{\xi}, \nabla \mathcal{H} \rangle = 0$). *However,* **gradient descent on the Hamiltonian converges to the Nash equilibrium.** *The remainder of the paper explores the implications and limitations of this insight.*

Papadimitriou & Piliouras (2016) recently analyzed the dynamics of Matching Pennies (essentially, the above example) and showed with some effort that the cyclic behavior covers the entire parameter space. The Hamiltonian reformulation directly explains the cyclic behavior via a conservation law.

### 2.2. The generalized Helmholtz decomposition

The **Hessian** of a game is the $(d \times d)$-matrix of second-derivatives $\mathbf{H}(\mathbf{w}) := \nabla_{\mathbf{w}} \cdot \boldsymbol{\xi}(\mathbf{w})^\intercal = \left( \frac{\partial \xi_\alpha(\mathbf{w})}{\partial w_\beta} \right)_{\alpha, \beta=1}^d$. Concretely, the Hessian can be written

$$\mathbf{H}(\mathbf{w}) = \begin{pmatrix} \nabla^2_{\mathbf{w}_1} \ell_1 & \nabla^2_{\mathbf{w}_1, \mathbf{w}_2} \ell_1 & \cdots & \nabla^2_{\mathbf{w}_1, \mathbf{w}_n} \ell_1 \\ \nabla^2_{\mathbf{w}_2, \mathbf{w}_1} \ell_2 & \nabla^2_{\mathbf{w}_2} \ell_2 & \cdots & \nabla^2_{\mathbf{w}_2, \mathbf{w}_n} \ell_2 \\ \vdots & & & \vdots \\ \nabla^2_{\mathbf{w}_n, \mathbf{w}_1} \ell_n & \nabla^2_{\mathbf{w}_n, \mathbf{w}_2} \ell_n & \cdots & \nabla^2_{\mathbf{w}_n} \ell_n \end{pmatrix},$$

where $\nabla^2_{\mathbf{w}_i, \mathbf{w}_j} \ell_k$ is the $(d_i \times d_j)$-block of $2^{\text{nd}}$-order derivatives. The Hessian of a game is not necessarily symmetric. Note: $\alpha, \beta$ run over dimensions; $i, j$ over players.

**Lemma 1** (generalized Helmholtz decomposition)**.** *The Hessian of any vector field decomposes uniquely into two components* $\mathbf{H}(\mathbf{w}) = \mathbf{S}(\mathbf{w}) + \mathbf{A}(\mathbf{w})$ *where* $\mathbf{S} \equiv \mathbf{S}^\intercal$ *is symmetric and* $\mathbf{A} + \mathbf{A}^\intercal \equiv 0$ *is antisymmetric.*

*Proof.* Any matrix decomposes uniquely as $\mathbf{M} = \mathbf{S} + \mathbf{A}$ where $\mathbf{S} = \frac{1}{2}(\mathbf{M} + \mathbf{M}^\intercal)$ and $\mathbf{A} = \frac{1}{2}(\mathbf{M} - \mathbf{M}^\intercal)$ are symmetric and antisymmetric. Applying the decomposition to the game Hessian yields the result. $\square$

The decomposition is preserved by orthogonal change-of-coordinates $\mathbf{P}^\intercal \mathbf{M} \mathbf{P} = \mathbf{P}^\intercal \mathbf{S} \mathbf{P} + \mathbf{P}^\intercal \mathbf{A} \mathbf{P}$ since the terms remain symmetric and antisymmetric.

The connection to the classical Helmholtz decomposition in calculus is sketched in appendix E. Two obvious classes of games arise from the decomposition:

**Definition 2.** *A game is a* ***potential game*** *if* $\mathbf{A}(\mathbf{w}) \equiv 0$. *It is a* ***Hamiltonian game*** *if* $\mathbf{S}(\mathbf{w}) \equiv 0$.

Potential games are well-studied and easy to solve. Hamiltonian games are a new class of games that are also easy to solve. The general case is more delicate, see section 3.

## 2.3. Stable fixed points

Gradient-based methods can reliably find local – but not global – optima of nonconvex objective functions (Lee et al., 2016; 2017). Similarly, gradient-based methods cannot be expected to find global Nash equilibria in games.

**Definition 3.** *A fixed point $\mathbf{w}^*$, with $\boldsymbol{\xi}(\mathbf{w}^*) = 0$, is* **stable** *if $\mathbf{S}(\mathbf{w}) \succeq 0$ and* **unstable** *if $\mathbf{S}(\mathbf{w}) \prec 0$ for $\mathbf{w}$ in a neighborhood of $\mathbf{w}^*$.*

Fixed points that are neither positive nor negative definite are beyond the scope of the paper.

**Lemma 2** (Stable fixed points are local Nash equilibria). *A point $\mathbf{w}^*$ is a* **local Nash equilibrium** *if, for all $i$, there is a neighborhood $U_i$ of $\mathbf{w}_i^*$ such that $\ell_i(\mathbf{w}_i, \mathbf{w}_{-i}^*) \geq \ell_i(\mathbf{w}_i^*, \mathbf{w}_{-i}^*)$ for $\mathbf{w}_i \in U_i$. If fixed point $\mathbf{w}^*$ is stable then it is a local Nash equilibrium.*

*Proof.* If $\mathbf{S}$ is positive semidefinite then so are its $(d_i \times d_i)$-submatrices $\mathbf{S}_i := \nabla^2_{\mathbf{w}_i} \ell_i$ for all $i$. The result follows. $\square$

Appendix A contains more details on local Nash equilibria.

## 2.4. Potential games

Potential games were introduced in Monderer & Shapley (1996). A game is a potential game if there is a single potential function $\phi : \mathbb{R}^d \to \mathbb{R}$ and positive numbers $\{\alpha_i > 0\}_{i=1}^n$ such that

$$\phi(\mathbf{w}_i', \mathbf{w}_{-i}) - \phi(\mathbf{w}_i'', \mathbf{w}_{-i}) = \alpha_i \Big( \ell_i(\mathbf{w}_i', \mathbf{w}_{-i}) - \ell_i(\mathbf{w}_i'', \mathbf{w}_{-i}) \Big)$$

for all $i$ and all $\mathbf{w}_i', \mathbf{w}_i'', \mathbf{w}_{-i}$. Monderer & Shapley (1996) show a game is a potential game iff $\alpha_i \nabla_{\mathbf{w}_i} \ell_i = \nabla_{\mathbf{w}_i} \phi$ for all $i$, which is equivalent to

$$\alpha_i \nabla^2_{\mathbf{w}_i \mathbf{w}_j} \ell_i = \alpha_j \nabla^2_{\mathbf{w}_i \mathbf{w}_j} \ell_j = \alpha_j \left( \nabla^2_{\mathbf{w}_j \mathbf{w}_i} \ell_j \right)^\intercal \quad \forall i, j.$$

Our definition of potential game is the special case where $\alpha_i = 1$ for all $i$, which Monderer & Shapley (1996) call an **exact** potential game. We use the shorthand 'potential game' to refer to exact potential games in what follows.

Potential games have been extensively studied since they are one of the few classes of games for which Nash equilibria can be computed (Rosenthal, 1973). For our purposes, they are games where simultaneous gradient descent on the losses is gradient descent on a single function. It follows that descent on $\boldsymbol{\xi}$ converges to a fixed point that is a local minimum of $\phi$ – or saddle, but see Lee et al. (2017).

## 2.5. Hamiltonian games

A concrete example may help understand antisymmetric matrices. Suppose $n$ competitors play one-on-one and that the probability of player $i$ beating player $j$ is $p_{ij}$. Then, assuming there are no draws, the probabilities satisfy $p_{ij} + p_{ji} = 1$ and $p_{ii} = \frac{1}{2}$. The matrix $\mathbf{A} = \left( \log \frac{p_{ij}}{1 - p_{ij}} \right)_{i,j=1}^n$ of logits is then antisymmetric. Intuitively, antisymmetry reflects a *hyperadversarial* setting where all pairwise interactions between players are zero-sum. In general, Hamiltonian games are related to – but distinct from – zero-sum games.

**Example 2** (an unconstrained[2] bimatrix game is zero–sum iff it is Hamiltonian). *Consider bimatrix game with $\ell_1(\mathbf{x}, \mathbf{y}) = \mathbf{x}^\intercal \mathbf{P} \mathbf{y}$ and $\ell_2(\mathbf{x}, \mathbf{y}) = \mathbf{x}^\intercal \mathbf{Q} \mathbf{y}$. Then $\boldsymbol{\xi} = (\mathbf{P}\mathbf{y}, \mathbf{Q}^\intercal \mathbf{x})$ and the Hessian components have block structure*

$$2\mathbf{A} = \begin{pmatrix} 0 & \mathbf{P} - \mathbf{Q} \\ (\mathbf{Q} - \mathbf{P})^\intercal & 0 \end{pmatrix} \quad 2\mathbf{S} = \begin{pmatrix} 0 & \mathbf{P} + \mathbf{Q} \\ (\mathbf{P} + \mathbf{Q})^\intercal & 0 \end{pmatrix}$$

*The game is Hamiltonian iff $\mathbf{S} = 0$ iff $\mathbf{P} + \mathbf{Q} = 0$ iff $\ell_1 + \ell_2 = 0$.*

There are Hamiltonian games that are not zero-sum and vice versa.

**Example 3** (Hamiltonian game that is not zero-sum). *Fix constants $a$ and $b$ and suppose players 1 and 2 minimize losses*

$$\ell_1(x, y) = x(y - b) \text{ and } \ell_2(x, y) = -(x - a)y$$

*with respect to $x$ and $y$ respectively.*

**Example 4** (zero-sum game that is not Hamiltonian). *Players 1 and 2 minimize*

$$\ell_1(x, y) = x^2 + y^2 \qquad \ell_2(x, y) = -(x^2 + y^2).$$

*The game actually has potential function $\phi(x, y) = x^2 - y^2$.*

Hamiltonian games are quite different from potential games. There is a Hamiltonian function $\mathcal{H}$ that specifies a conserved quantity. Whereas the dynamics *equal* $\nabla \phi$ in potential games; they are *orthogonal* to $\nabla \mathcal{H}$ in Hamiltonian games. The orthogonality implies the conservation law that underlies the cyclic behavior in example 1.

**Theorem 3.** *Let $\mathcal{H}(\mathbf{w}) := \frac{1}{2} \|\boldsymbol{\xi}(\mathbf{w})\|_2^2$. If the game is Hamiltonian then (i) $\nabla \mathcal{H} = \mathbf{A}^\intercal \boldsymbol{\xi}$ and (ii) $\boldsymbol{\xi}$ preserves the level sets of $\mathcal{H}$ since $\langle \boldsymbol{\xi}, \nabla \mathcal{H} \rangle = 0$. If the Hessian is invertible and $\lim_{\|\mathbf{w}\| \to \infty} \mathcal{H}(\mathbf{w}) = \infty$ then (iii) gradient descent on $\mathcal{H}$ converges to a local Nash equilibrium.*

In fact, $\mathcal{H}$ is a Hamiltonian[3] function for the game dynamics. We use the notation $\mathcal{H}(\mathbf{w}) = \frac{1}{2} \|\boldsymbol{\xi}(\mathbf{w})\|^2$ throughout the paper. However, $\mathcal{H}$ is only a Hamiltonian function for $\boldsymbol{\xi}$ if the game is Hamiltonian.

---

[2]The parameters in bimatrix games are usually constrained to the probability simplex.

[3]Notation: $\mathcal{H}$ for Hamiltonian; $\mathbf{H}$ for Hessian.

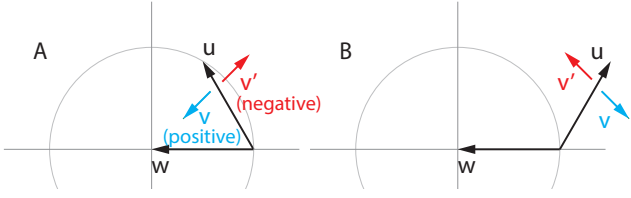

*Figure 2. Infinitesimal alignment* is positive (cyan) when small positive $\lambda$ either: **(A)** pulls **u** toward **w**, if **w** and **u** have angle $< 90°$; or **(B)** pushes **u** away from **w** if their angle is $> 90°$.

There is a precise mapping from Hamiltonian games to symplectic geometry, see Appendix E. Symplectic geometry is the modern formulation of classical mechanics. Recall that periodic behaviors (e.g. orbits) often arise in classical mechanics. The orbits lie on the level sets of the Hamiltonian, which expresses the total energy of the system.

## 3. Algorithms

Fixed points of potential and Hamiltonian games can be found by descent on $\boldsymbol{\xi}$ and $\nabla \mathcal{H}$ respectively. This section tackles finding stable fixed points in general games.

### 3.1. Finding fixed points

If the Hessian $\mathbf{H}(\mathbf{w})$ is invertible, then $\nabla \mathcal{H} = \mathbf{H}^\mathsf{T} \boldsymbol{\xi} = 0$ iff $\boldsymbol{\xi} = 0$. Thus, gradient descent on $\mathcal{H}$ converges to fixed points of $\boldsymbol{\xi}$. However, there is no guarantee that descent on $\mathcal{H}$ will find a *stable* fixed point.

Mescheder et al. (2017) propose *consensus optimization*, a gradient adjustment of the form

$$\boldsymbol{\xi} + \lambda \cdot \mathbf{H}^\mathsf{T} \boldsymbol{\xi} = \boldsymbol{\xi} + \lambda \cdot \nabla \mathcal{H}.$$

Unfortunately, consensus optimization can converge to unstable fixed points even in simple cases where the 'game' is to minimize a single function:

**Example 5** (consensus optimization converges to global maximum). *Consider a potential game with losses $\ell_1(x,y) = \ell_2(x,y) = -\frac{\kappa}{2}(x^2 + y^2)$ with $\kappa \gg 0$. Then*

$$\boldsymbol{\xi} = -\kappa \cdot \begin{pmatrix} x \\ y \end{pmatrix} \text{ and } \mathbf{H} = -\begin{pmatrix} \kappa & 0 \\ 0 & \kappa \end{pmatrix}$$

*Note that $\|\boldsymbol{\xi}\|^2 = \kappa^2(x^2 + y^2)$ and*

$$\boldsymbol{\xi} + \lambda \cdot \mathbf{H}^\mathsf{T} \boldsymbol{\xi} = \kappa(\lambda\kappa - 1) \cdot \begin{pmatrix} x \\ y \end{pmatrix}.$$

*Descent on $\boldsymbol{\xi} + \lambda \cdot \mathbf{H}^\mathsf{T} \boldsymbol{\xi}$ converges to the global maximum $(x,y) = (0,0)$ unless $\lambda < \frac{1}{\kappa}$.*

Although consensus optimization works well in some special cases like two-player zero-sum; it cannot be considered a candidate algorithm for finding stable fixed points in general games, since it fails in the basic case of potential games.

---

**Algorithm 1** Symplectic Gradient Adjustment

**Input:** losses $\mathcal{L} = \{\ell_i\}_{i=1}^n$, weights $\mathcal{W} = \{\mathbf{w}_i\}_{i=1}^n$
  $\boldsymbol{\xi} \leftarrow$ `[gradient(`$\ell_i$`,` $\mathbf{w}_i$`) for` $(\ell_i, \mathbf{w}_i) \in (\mathcal{L}, \mathcal{W})$`]`
  $\mathbf{A}^\mathsf{T}\boldsymbol{\xi} \leftarrow$ `get_sym_adj(`$\mathcal{L}, \mathcal{W}$`)`     *// appendix C*
  **if** `align` **then**
    $\nabla\mathcal{H} \leftarrow$ `[gradient(`$\frac{1}{2}\|\boldsymbol{\xi}\|^2$`,` $\mathbf{w}$`) for` $\mathbf{w} \in \mathcal{W}$`]`
    $\lambda \leftarrow$ `sign`$\left(\frac{1}{d}\langle\boldsymbol{\xi}, \nabla\mathcal{H}\rangle\langle\mathbf{A}^\mathsf{T}\boldsymbol{\xi}, \nabla\mathcal{H}\rangle + \epsilon\right)$   *// $\epsilon = \frac{1}{10}$*
  **else**
    $\lambda \leftarrow 1$
  **end if**
**Output:** $\boldsymbol{\xi} + \lambda \cdot \mathbf{A}^\mathsf{T}\boldsymbol{\xi}$     *// plug into any optimizer*

---

### 3.2. Finding stable fixed points

There are two classes of games where we know how to find stable fixed points: potential games where $\boldsymbol{\xi}$ converges to a local minimum and Hamiltonian games where $\nabla\mathcal{H}$, which is orthogonal to $\boldsymbol{\xi}$, finds stable fixed points. In the general case, the following desiderata provide a set of reasonable properties for an adjustment $\boldsymbol{\xi}_\lambda$ of the game dynamics:

**Desiderata.** *To find stable fixed points, an adjustment $\boldsymbol{\xi}_\lambda$ to the game dynamics should satisfy*

*D1.* compatible with game dynamics: $\langle\boldsymbol{\xi}_\lambda, \boldsymbol{\xi}\rangle = \alpha_1 \cdot \|\boldsymbol{\xi}\|^2$;

*D2.* compatible with potential dynamics: *if the game is a potential game then* $\langle\boldsymbol{\xi}_\lambda, \nabla\phi\rangle = \alpha_2 \cdot \|\nabla\phi\|^2$;

*D3.* compatible with Hamiltonian dynamics: *If the game is Hamiltonian then* $\langle\boldsymbol{\xi}_\lambda, \nabla\mathcal{H}\rangle = \alpha_3 \cdot \|\nabla\mathcal{H}\|^2$;

*D4.* attracted to stable equilibria: *in neighborhoods where* $\mathbf{S} \succ 0$, *require* $\theta(\boldsymbol{\xi}_\lambda, \nabla\mathcal{H}) \leq \theta(\boldsymbol{\xi}, \nabla\mathcal{H})$;

*D5.* repelled by unstable equilibria: *in neighborhoods where* $\mathbf{S} \prec 0$, *require* $\theta(\boldsymbol{\xi}_\lambda, \nabla\mathcal{H}) \geq \theta(\boldsymbol{\xi}, \nabla\mathcal{H})$;

*for some $\alpha_1, \alpha_2, \alpha_3 > 0$.*

Desideratum *D1* does not guarantee that players act in their own self-interest – this requires a stronger positivity condition on dot-products with subvectors of $\boldsymbol{\xi}$, see Balduzzi (2017). Desiderata *D4* and *D5* are explained in section 3.4. The unadjusted dynamics $\boldsymbol{\xi}$ satisfies all the desiderata except *D3*. Consensus optimization only satisfies *D3* and *D4*. Ideally, desideratum *D5* would be strengthened to 'repelled by saddle points where $\boldsymbol{\xi}(\mathbf{w}) = 0$ but $\mathbf{S}(\mathbf{w}) \not\succeq 0$'.

### 3.3. Symplectic gradient adjustment

**Proposition 4.** *The **symplectic gradient adjustment (SGA)***

$$\boldsymbol{\xi}_\lambda := \boldsymbol{\xi} + \lambda \cdot \mathbf{A}^\mathsf{T}\boldsymbol{\xi}.$$

*satisfies D1–D3 for $\lambda > 0$, with $\alpha_1 = 1 = \alpha_2$ and $\alpha_3 = \lambda$.*

Note that desiderata $D1$ and $D2$ are true even when $\lambda < 0$. This will prove useful, since example 6 and theorem 5 show it is necessary pick negative $\lambda$ near $\mathbf{S} \prec 0$. Section 3.4 shows how to also satisfy desiderata $D4$ and $D5$.

*Proof.* First claim: $\lambda \cdot \boldsymbol{\xi}^\mathsf{T} \mathbf{A}^\mathsf{T} \boldsymbol{\xi} = 0$ by skew-symmetry of $\mathbf{A}$. Second claim: $\mathbf{A} \equiv 0$ in a potential game, so $\boldsymbol{\xi}_\lambda = \boldsymbol{\xi} = \nabla\phi$. Third claim: $\langle \boldsymbol{\xi}_\lambda, \nabla\mathcal{H} \rangle = \langle \boldsymbol{\xi}_\lambda, \mathbf{H}^\mathsf{T} \boldsymbol{\xi} \rangle = \langle \boldsymbol{\xi}, \mathbf{A}^\mathsf{T} \boldsymbol{\xi} \rangle = \lambda \cdot \boldsymbol{\xi}^\mathsf{T} \mathbf{A} \mathbf{A}^\mathsf{T} \boldsymbol{\xi} = \lambda \cdot \|\nabla\mathcal{H}\|^2$ since $\mathbf{H} = \mathbf{A}$ by assumption and since $\boldsymbol{\xi}^\mathsf{T} \mathbf{A}^\mathsf{T} \boldsymbol{\xi} = 0$ by antisymmetry. □

In the example below, almost any choice of positive $\lambda$ results in convergence to an unstable equilibrium. The problem arises from the combination of a weak repellor with a strong rotational force. The next section shows how to pick $\lambda$ to avoid unstable equilibria.

**Example 6** (failure case for $\lambda > 0$). *Suppose $\epsilon > 0$ is small and*

$$f(x,y) = -\frac{\epsilon}{2}x^2 - xy \text{ and } g(x,y) = -\frac{\epsilon}{2}y^2 + xy$$

*with an unstable equilibrium at $(0,0)$. The dynamics are*

$$\boldsymbol{\xi} = \epsilon \cdot \begin{pmatrix} -x \\ -y \end{pmatrix} + \begin{pmatrix} -y \\ x \end{pmatrix} \quad \text{with} \quad \mathbf{A} = \begin{pmatrix} 0 & -1 \\ 1 & 0 \end{pmatrix}$$

*and*

$$\mathbf{A}^\mathsf{T} \boldsymbol{\xi} = \begin{pmatrix} x \\ y \end{pmatrix} + \epsilon \begin{pmatrix} -y \\ x \end{pmatrix}$$

*Finally observe that*

$$\boldsymbol{\xi} + \lambda \cdot \mathbf{A}^\mathsf{T} \boldsymbol{\xi} = (\lambda - \epsilon) \cdot \begin{pmatrix} x \\ y \end{pmatrix} + (1 + \epsilon\lambda) \cdot \begin{pmatrix} -y \\ x \end{pmatrix}$$

*which converges to the unstable equilibrium if $\lambda > \epsilon$.*

Lemma 9 in the appendix shows that, if $\mathbf{S}$ and $\mathbf{A}$ commute and $\mathbf{S} \succeq 0$, then $\langle \boldsymbol{\xi}_\lambda, \nabla\mathcal{H} \rangle \geq 0$ for $\lambda > 0$. The proof of theorem 5 introduces the additive condition number to upper-bound the worst-case noncommutativity of $\mathbf{S}$, which allows to quantify the relationship between $\boldsymbol{\xi}_\lambda$ and $\nabla\mathcal{H}$.

**Theorem 5.** *Let $\mathbf{S}$ be a symmetric matrix with eigenvalues $\sigma_{max} \geq \cdots \geq \sigma_{min}$. The **additive condition number**[4] of $\mathbf{S}$ is $\kappa = \sigma_{max} - \sigma_{min}$. If $\mathbf{S} \succeq 0$ is positive semidefinite with additive condition number $\kappa$ then $\lambda \in (0, \frac{4}{\kappa})$ implies*

$$\langle \boldsymbol{\xi}_\lambda, \nabla\mathcal{H} \rangle \geq 0.$$

*If $\mathbf{S}$ is negative semidefinite, then $\lambda \in (0, \frac{4}{\kappa})$ implies*

$$\langle \boldsymbol{\xi}_{-\lambda}, \nabla\mathcal{H} \rangle \leq 0.$$

*The inequalities are strict if $\mathbf{H}$ is invertible.*

If $\kappa = 0$, then $\mathbf{S} = \sigma \cdot \mathbf{I}$ commutes with ***all*** matrices. The larger the additive condition number $\kappa$, the larger the *potential* failure of $\mathbf{S}$ to commute with other matrices.

---

[4]The condition number of a positive definite matrix is $\frac{\sigma_{max}}{\sigma_{min}}$.

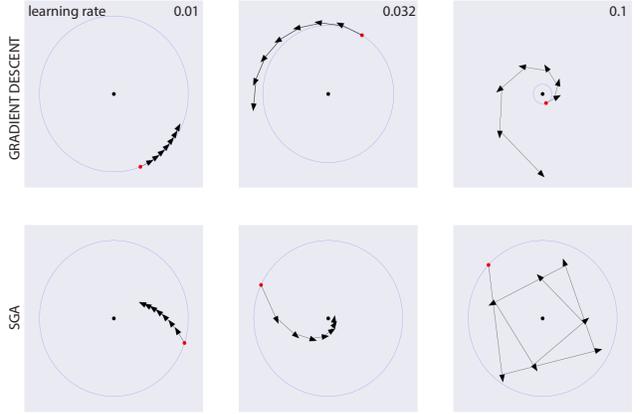

**learning rate**

GRADIENT DESCENT    0.01    0.032    0.1

SGA

*Figure 3.* SGA allows faster and more robust convergence to stable fixed points. Note the scale of top-right panel differs from rest.

### 3.4. How to pick $\mathrm{sign}(\lambda)$

This section explains desiderata $D4$–$D5$ and shows how to pick $\mathrm{sign}(\lambda)$ to speed up convergence towards stable and away from unstable fixed points.

First, observe that $\langle \boldsymbol{\xi}, \nabla\mathcal{H} \rangle = \boldsymbol{\xi}^\mathsf{T}(\mathbf{S} + \mathbf{A})^\mathsf{T} \boldsymbol{\xi} = \boldsymbol{\xi}^\mathsf{T} \mathbf{S} \boldsymbol{\xi}$. It follows that for $\boldsymbol{\xi} \neq 0$:

$$\begin{cases} \text{if } \mathbf{S} \succeq 0 & \text{then } \langle \boldsymbol{\xi}, \nabla\mathcal{H} \rangle \geq 0; \\ \text{if } \mathbf{S} \prec 0 & \text{then } \langle \boldsymbol{\xi}, \nabla\mathcal{H} \rangle < 0. \end{cases} \quad (1)$$

A criterion to probe the positive/negative definiteness of $\mathbf{S}$ is thus to check the sign of $\langle \boldsymbol{\xi}, \nabla\mathcal{H} \rangle$. The dot product can take any value if $\mathbf{S}$ is not positive nor negative definite. The behavior near saddles is beyond the scope of the paper.

Desiderata $D4$ can be interpreted as follows. If $\boldsymbol{\xi}$ points at a stable equilibrium then we require that $\boldsymbol{\xi}_\lambda$ points *more* towards the equilibrium (i.e. has smaller angle). Conversely if $\boldsymbol{\xi}$ points away then the adjustment should point *further* away. More formally,

**Definition 4.** *Let $\mathbf{u}$ and $\mathbf{v}$ be two vectors. The **infinitesimal alignment** of $\boldsymbol{\xi}_\lambda := \mathbf{u} + \lambda \cdot \mathbf{v}$ with a third vector $\mathbf{w}$ is*

$$\mathrm{align}(\boldsymbol{\xi}_\lambda, \mathbf{w}) := \frac{d}{d\lambda}\left\{ \cos^2 \theta_\lambda \right\}_{|\lambda=0} \text{ for } \theta_\lambda := \theta(\boldsymbol{\xi}_\lambda, \mathbf{w}).$$

If $\mathbf{u}$ and $\mathbf{w}$ point the same way, $\mathbf{u}^\mathsf{T} \mathbf{w} > 0$, then $\mathrm{align} > 0$ when $\mathbf{v}$ bends $\mathbf{u}$ further toward $\mathbf{w}$, see figure 2A. Otherwise $\mathrm{align} > 0$ when $\mathbf{v}$ bends $\mathbf{u}$ away from $\mathbf{w}$, figure 2B.

**Proposition 6.** *Desiderata $D4$–$D5$ are satisfied for $\lambda$ such that $\lambda \cdot \langle \boldsymbol{\xi}, \nabla\mathcal{H} \rangle \cdot \langle \mathbf{A}^\mathsf{T} \boldsymbol{\xi}, \nabla\mathcal{H} \rangle \geq 0$.*

Computing the sign of $\langle \boldsymbol{\xi}, \nabla\mathcal{H} \rangle$ provides a check for stable and unstable fixed points. Computing the sign of $\langle \mathbf{A}^\mathsf{T} \boldsymbol{\xi}, \nabla\mathcal{H} \rangle$ checks whether the adjustment term points towards or away from the (nearby) fixed point. Putting the two checks together yields a prescription for the sign of $\lambda$.

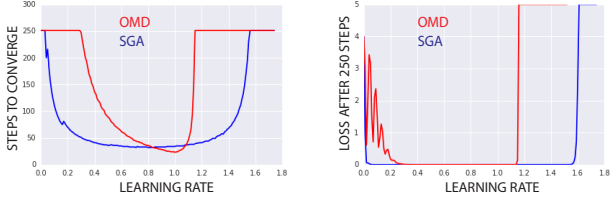

Figure 4. Comparison of SGA with optimistic mirror descent. Sweep over learning rates in range $[0.01, 1.75]$; $\lambda = 1$ throughout. **(Left):** iterations to convergence (up to 250). **(Right):** average absolute value of losses over iterations 240-250, cutoff at 5.

**Alignment and convergence rates.** Finally, we show that increasing alignment helps speed convergence. Gradient descent is also known as the method of steepest descent. In general games, however, $\xi$ does not follow the steepest descent path to fixed points due to the 'rotational force', which forces lower learning rates and slows down convergence:

**Theorem 7.** *Suppose $f$ is convex and Lipschitz smooth with $\|\nabla f(\mathbf{x}) - \nabla f(\mathbf{y})\| \leq L \cdot \|\mathbf{x} - \mathbf{y}\|$. Let $\mathbf{w}_{t+1} = \mathbf{w}_t - \eta \cdot \mathbf{v}$ where $\|\mathbf{v}\| = \|\nabla f(\mathbf{w}_t)\|$. Then the optimal step size is $\eta^* = \frac{\cos \theta}{L}$ where $\theta := \theta(\nabla f(\mathbf{w}_t), \mathbf{v})$, with*

$$f(\mathbf{w}_{t+1}) \leq f(\mathbf{w}_t) - \frac{\cos^2 \theta}{2L} \cdot \|\nabla f(\mathbf{w}_t)\|^2.$$

Increasing the cosine with the steepest direction improves convergence. The alignment computation in algorithm 1 chooses $\lambda$ to be positive or negative such that $\xi_\lambda$ is bent towards stable (increasing the cosine) and away from unstable fixed points. Adding a small $\epsilon > 0$ to the computation introduces a weak bias towards stable fixed points.

### 3.5. Aligned consensus optimization

The stability criterion in (1) also provides a simple way to prevent consensus optimization from converging to unstable equilibria. **Aligned consensus optimization** is

$$\xi + |\lambda| \cdot \text{sign}\left(\langle \xi, \nabla \mathcal{H} \rangle\right) \cdot \mathbf{H}^\mathsf{T} \xi, \quad (2)$$

where in practice we set $\lambda = 1$. Aligned consensus satisfies desiderata *D3–D5*. However, it behaves strangely in potential games. Multiplying by the Hessian is the 'inverse' of Newton's method: it increases the gap between small and large eigenvalues, increasing the (usual, multiplicative) condition number and slows down convergence. Nevertheless, consensus optimization works well in GANs (Mescheder et al., 2017), and aligned consensus may improve performance, see figure 9 in appendix.

Finally, note that dropping the first term $\xi$ from (2) yields a simpler update that also satisfies *D3–D5*. However, the resulting algorithm performs poorly in experiments (not shown), perhaps because it is attracted to saddles.

## 4. Experiments

We compare SGA with simultaneous gradient descent, optimistic mirror descent (Daskalakis et al., 2018) and consensus optimization (Mescheder et al., 2017) in basic settings.

### 4.1. Learning rates and alignment

We investigate the effect of SGA when a weak attractor is coupled to a strong rotational force:

$$\ell_1(x, y) = \frac{1}{2}x^2 + 10xy \quad \text{and} \quad \ell_2(x, y) = \frac{1}{2}y^2 - 10xy$$

Gradient descent is extremely sensitive to the choice of learning rate $\eta$, top row of figure 3. As $\eta$ increases through $\{0.01, 0.032, 0.1\}$ gradient descent goes from converging extremely slowly, to diverging slowly, to diverging rapidly. SGA yields faster, more robust convergence. SGA converges faster with learning rates $\eta = 0.01$ and $\eta = 0.032$, and only starts overshooting the fixed point for $\eta = 0.1$.

### 4.2. Basic adversarial games

Figure 4 compares SGA with optimistic mirror descent on a zero-sum bimatrix game with $\ell_{1/2}(\mathbf{w}_1, \mathbf{w}_2) = \pm \mathbf{w}_1^\mathsf{T} \mathbf{w}_2$. The example is modified from Daskalakis et al. (2018) who also consider a linear offset that makes no difference. A run converges, panel A, if the average absolute value of losses on the last 10 iterations is $< 0.01$.

Although OMD's peak performance is better than SGA, we find that SGA converges – and does so faster – for a wider range of learning rates. OMD diverges for learning rates not in the range $[0.3, 1.2]$. Simultaneous gradient descent oscillates without converging (not shown). Individual runs are shown in the appendix. The appendix also compares the performance of the algorithms on four-player games.

### 4.3. Generative adversarial networks

We apply SGA to a basic setup adapted from Metz et al. (2017). Data is sampled from a highly multimodal distribution designed to probe the tendency to collapse onto a subset of modes during training. The distribution is a mixture of 16 Gaussians arranged in a $4 \times 4$ grid, see ground truth in figure 8. The generator and discriminator networks both have 6 ReLU layers of 384 neurons. The generator has two output neurons; the discriminator has one.

Figure 5 shows results after $\{2000, 4000, 6000, 8000\}$ iterations. The networks are trained under RMSProp. Learning rates were chosen by visual inspection of grid search results at iteration 8000, see appendix. Simultaneous gradient descent and SGA are shown in the figure. Results for consensus optimization are in the appendix.

Simultaneous gradient descent exhibits mode collapse fol-lowed by mode hopping in later iterations (not shown). Mode hopping is analogous to the cycles in example 1. Unaligned SGA converges to the correct distribution; alignment speeds up convergence slightly. Consensus optimization performs similarly, see figure 9 in the appendix. However, it can converge to local maxima, recall example 5.

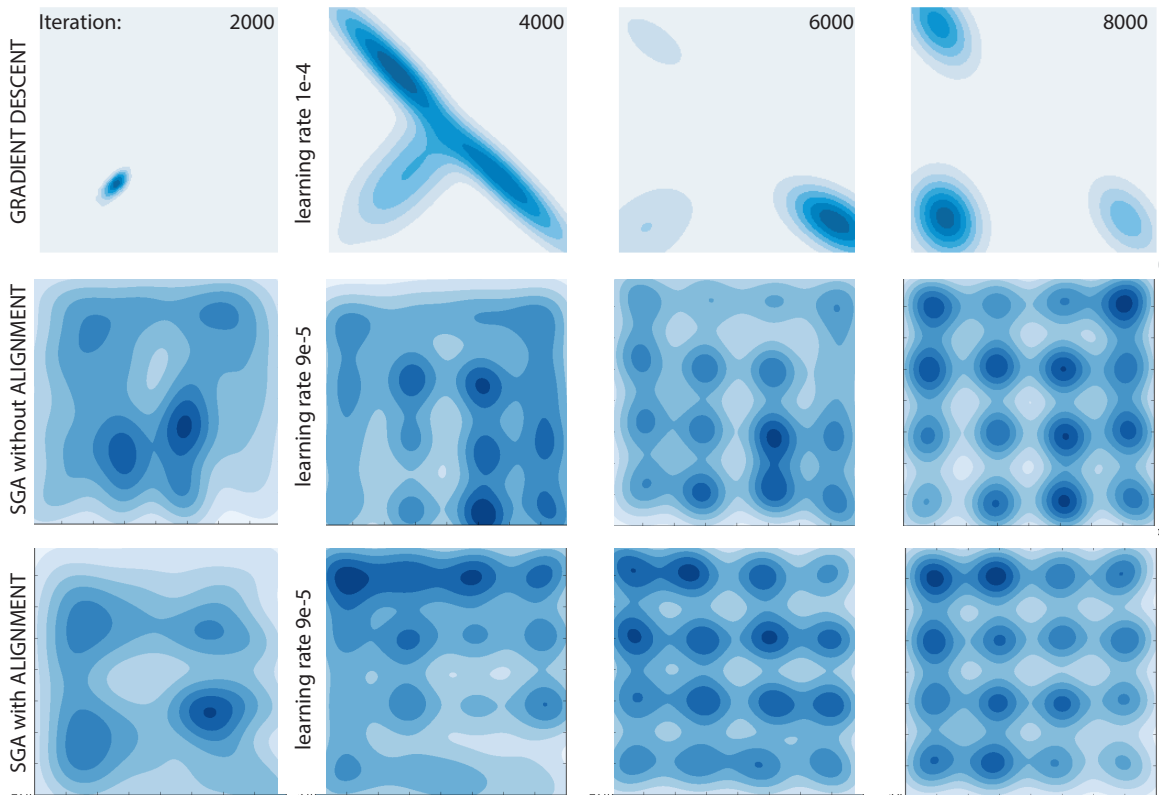

*Figure 5.* **Top:** Simultaneous gradient descent suffers from mode collapse and in later iterations (not shown) mode hopping. **Middle:** vanilla SGA converges smoothly to the ground truth (figure 8 in appendix). **Bottom:** SGA with alignment converges slightly faster.

## 5. Discussion

Modern deep learning treats differentiable modules like plug-and-play lego blocks. For this to work, at the very least, we need to know that gradient descent will find local minima. Unfortunately, gradient descent does *not* necessarily find local minima when optimizing multiple interacting objectives. With the recent proliferation of algorithms that optimize more than one loss, it is becoming increasingly urgent to understand and control the dynamics of interacting losses. Although there is interesting recent work on two-player adversarial games such as GANs, there is essentially no work on finding stable fixed points in general.

The generalized Helmholtz decomposition provides a powerful new perspective on game dynamics. A key feature is that the analysis is indifferent to the number of players. Instead, it is the interplay between the simultaneous gradient $\xi$ on the losses and the symmetric and antisymmetric matrices of second-order terms that guides algorithm design and governs the dynamics under gradient adjustments.

Symplectic gradient adjustment is a straightforward application of the Helmholtz decomposition. It is unlikely that SGA is the only or best approach to finding stable fixed points. A deeper understanding of the interaction between the potential and Hamiltonian components will lead to more effective algorithms. Of particular interest are pure first-order methods that do not use Hessian-vector products.

Finally, it is worth raising a philosophical point. The goal in this paper is to find stable fixed points (e.g. because in GANs they yield pleasing samples). We are not concerned with the losses of the players *per se*. The gradient adjustments may lead to a player acting against its own self-interest by increasing its loss. We consider this acceptable insofar as it encourages convergence to a stable fixed point.

**Acknowledgements.** We thank Guillame Desjardins, Csaba Szepesvari and especially Alistair Letcher for useful feedback.

# APPENDIX

## A. Stable fixed points vs local Nash equilibria

The main text introduces two solutions concepts for general games: stable fixed points and local Nash equilibria. Lemma 2 shows that if $\mathbf{w}^*$ is a stable fixed point then it is also a local Nash equilibrium. If the game is two-player zero-sum, then the converse also holds:

**Lemma 8** (local Nash equilibria are stable fixed points in two-player zero-sum games). *If a game is two-player zero-sum, then local Nash equilibria are stable fixed points.*

*Proof.* Consider a two-player zero-sum game with losses

$$f(\mathbf{x}, \mathbf{y}) \quad \text{and} \quad g(\mathbf{x}, \mathbf{y}) = -f(\mathbf{x}, \mathbf{y}).$$

Then $\boldsymbol{\xi} = (\nabla_{\mathbf{x}} f, -\nabla_{\mathbf{y}} f)$ and

$$
\begin{aligned}
\mathbf{H} &= \begin{pmatrix} \nabla_{\mathbf{x},\mathbf{x}}^2 f & \nabla_{\mathbf{x},\mathbf{y}}^2 f \\ -\nabla_{\mathbf{y},\mathbf{x}}^2 f & -\nabla_{\mathbf{y},\mathbf{y}}^2 f \end{pmatrix} \\
&= \underbrace{\begin{pmatrix} \nabla_{\mathbf{x},\mathbf{x}}^2 f & \mathbf{0} \\ \mathbf{0} & -\nabla_{\mathbf{y},\mathbf{y}}^2 f \end{pmatrix}}_{\mathbf{S}} + \underbrace{\begin{pmatrix} \mathbf{0} & \nabla_{\mathbf{x},\mathbf{y}}^2 f \\ -\nabla_{\mathbf{y},\mathbf{x}}^2 f & \mathbf{0} \end{pmatrix}}_{\mathbf{A}}
\end{aligned}
$$

Thus, $\mathbf{S}$ is positive semidefinite iff the blocks $\nabla_{\mathbf{x},\mathbf{x}}^2 f$ and $\nabla_{\mathbf{y},\mathbf{y}}^2 g = -\nabla_{\mathbf{y},\mathbf{y}}^2 f$ are both positive semidefinite. The result follows. □

The following example, due to Alistair Letcher, shows there are local Nash equilibria that are not stable fixed points:

**Example 7** (a local Nash equilibrium that is not stable). *Let*

$$\ell_1(x, y) = \frac{x^2}{2} + 2xy \quad \text{and} \quad \ell_2(x, y) = \frac{y^2}{2} + 2xy$$

*Then*

$$\boldsymbol{\xi} = \begin{pmatrix} x + 2y \\ y + 2x \end{pmatrix} \quad \text{and} \quad \mathbf{H} = \begin{pmatrix} 1 & 2 \\ 2 & 1 \end{pmatrix}.$$

*There is a fixed point at $(x, y) = (0, 0)$ which is a local Nash equilibrium since $\frac{\partial^2}{\partial x^2} \ell_1(0, 0) = 1 = \frac{\partial^2}{\partial y^2} \ell_2(0, 0) > 0$. However, $\mathbf{S} = \mathbf{H}$ has eigenvalues $\lambda_1 = 3$ and $\lambda_2 = -1$, and so is not positive semidefinite.*

**Which is the 'right' solution concept?** We prefer stable fixed points. In the case of GANs (two-player zero-sum) there is no difference between the solution concepts by lemma 8. However, the solution concepts differ for potential games. Example 7 is a potential game since $\mathbf{S} = \mathbf{H}$. It is easy to check that the potential function

$$\phi(x, y) = \frac{x^2}{2} + 2xy + \frac{y^2}{2}$$

satisfies $\boldsymbol{\xi} = \nabla \phi$ and $\mathbf{H} = \nabla^2 \phi$. The Nash equilibrium is a saddle-point for the potential function. An algorithm that is guaranteed to converge to the Nash equilibrium in example 7 is thus guaranteed to converge to the saddle point of the potential function $\phi$.

This rules out local Nash equilibrium as a solution concept for our purposes – since desideratum $D2$ is that the gradient adjustment behaves well for potential games and converging to saddles of the potential function does not count as good behavior.

## B. Proofs

**Proof of theorem 3.**

*Proof.* Direct computation shows $\nabla \mathcal{H} = \mathbf{H}^\mathsf{T} \boldsymbol{\xi}$ for any game. The first statement follows since $\mathbf{H} = \mathbf{A}$ in Hamiltonian games. For the second statement, the directional derivative is $D_{\boldsymbol{\xi}} \mathcal{H} = \langle \boldsymbol{\xi}, \nabla \mathcal{H} \rangle = \boldsymbol{\xi}^\mathsf{T} \mathbf{A}^\mathsf{T} \boldsymbol{\xi}$ where $\boldsymbol{\xi}^\mathsf{T} \mathbf{A}^\mathsf{T} \boldsymbol{\xi} = (\boldsymbol{\xi}^\mathsf{T} \mathbf{A}^\mathsf{T} \boldsymbol{\xi})^\mathsf{T} = \boldsymbol{\xi}^\mathsf{T} \mathbf{A} \boldsymbol{\xi} = -(\boldsymbol{\xi}^\mathsf{T} \mathbf{A}^\mathsf{T} \boldsymbol{\xi})$ since $\mathbf{A} = -\mathbf{A}^\mathsf{T}$ by skew-symmetry. It follows that $\boldsymbol{\xi}^\mathsf{T} \mathbf{A}^\mathsf{T} \boldsymbol{\xi} = 0$.

For the third, gradient descent on $\mathcal{H}$ will converge to a point where $\nabla \mathcal{H} = \mathbf{H}^\mathsf{T} \boldsymbol{\xi}(\mathbf{w}) = 0$. If the Hessian is invertible then clearly $\boldsymbol{\xi}(\mathbf{w}) = 0$. The fixed-point is a local Nash equilibrium by Lemma 2 since $0 \equiv \mathbf{S} \succeq 0$ in a Hamiltonian game. □

**Proof of theorem 5.** The following lemma, whilst not used in the proof of theorem 5, nevertheless provides a useful intuition about the role of commutativity.

Recall that two matrices $\mathbf{A}$ and $\mathbf{S}$ *commute* iff $[\mathbf{A}, \mathbf{S}] := \mathbf{AS} - \mathbf{SA} = \mathbf{0}$. That is, iff $\mathbf{AS} = \mathbf{SA}$.

**Lemma 9.** *If $\mathbf{S} \succeq 0$ is symmetric positive semidefinite and $\mathbf{S}$ commutes with $\mathbf{A}$ then*

$$\langle \boldsymbol{\xi}_\lambda, \nabla \mathcal{H} \rangle \geq 0 \text{ for all } \lambda \geq 0.$$

*Proof.* First observe that $\boldsymbol{\xi}^\intercal \mathbf{AS} \boldsymbol{\xi} = \boldsymbol{\xi}^\intercal \mathbf{S}^\intercal \mathbf{A}^\intercal \boldsymbol{\xi} = -\boldsymbol{\xi}^\intercal \mathbf{SA} \boldsymbol{\xi}$, where the first equality holds since the expression is a scalar, and the second holds since $\mathbf{S} = \mathbf{S}^\intercal$ and $\mathbf{A} = -\mathbf{A}^\intercal$. It follows that $\boldsymbol{\xi}^\intercal \mathbf{AS} \boldsymbol{\xi} = 0$ if $\mathbf{SA} = \mathbf{AS}$.

Rewrite the inequality as

$$\langle \boldsymbol{\xi}_\lambda \cdot \mathbf{A}^\intercal \boldsymbol{\xi}, \nabla \mathcal{H} \rangle = \langle \boldsymbol{\xi} + \lambda \cdot \mathbf{A}^\intercal \boldsymbol{\xi}, \mathbf{S} \boldsymbol{\xi} + \mathbf{A}^\intercal \boldsymbol{\xi} \rangle$$
$$= \boldsymbol{\xi}^\intercal \mathbf{S} \boldsymbol{\xi} + \lambda \boldsymbol{\xi}^\intercal \mathbf{AA}^\intercal \boldsymbol{\xi} \geq 0$$

since $\boldsymbol{\xi}^\intercal \mathbf{AS} \boldsymbol{\xi} = 0$ and by positivity of $\mathbf{S}$, $\lambda$ and $\mathbf{AA}^\intercal$. $\quad\square$

The proof of the theorem follows.

*Proof.* We prove the case $\mathbf{S} \succeq 0$; the case $\mathbf{S} \preceq 0$ is similar. Rewrite the inequality as

$$\langle \boldsymbol{\xi} + \lambda \cdot \mathbf{A}^\intercal \boldsymbol{\xi}, \nabla \mathcal{H} \rangle = (\boldsymbol{\xi} + \lambda \cdot \mathbf{A}^\intercal \boldsymbol{\xi})^\intercal \cdot (\mathbf{S} + \mathbf{A}^\intercal) \boldsymbol{\xi}$$
$$= \boldsymbol{\xi}^\intercal \mathbf{S} \boldsymbol{\xi} + \lambda \boldsymbol{\xi}^\intercal \mathbf{AS} \boldsymbol{\xi} + \lambda \boldsymbol{\xi}^\intercal \mathbf{AA}^\intercal \boldsymbol{\xi}$$

Let $\beta = \|A^\intercal \boldsymbol{\xi}\|$ and $\tilde{\mathbf{S}} = \mathbf{S} - \sigma_{\min} \cdot \mathbf{I}$, where $\mathbf{I}$ is the identity matrix. Then

$$\boldsymbol{\xi}^\intercal \mathbf{S} \boldsymbol{\xi} + \lambda \boldsymbol{\xi}^\intercal \mathbf{AS} \boldsymbol{\xi} + \lambda \cdot \beta^2 \geq \boldsymbol{\xi}^\intercal \tilde{\mathbf{S}} \boldsymbol{\xi} + \lambda \boldsymbol{\xi}^\intercal \mathbf{A} \tilde{\mathbf{S}} \boldsymbol{\xi} + \lambda \cdot \beta^2$$

since $\boldsymbol{\xi}^\intercal \mathbf{S} \boldsymbol{\xi} \geq \boldsymbol{\xi}^\intercal \tilde{\mathbf{S}} \boldsymbol{\xi}$ by construction and $\boldsymbol{\xi}^\intercal \mathbf{A} \tilde{\mathbf{S}} \boldsymbol{\xi} = \boldsymbol{\xi}^\intercal \mathbf{AS} \boldsymbol{\xi} - \sigma_{\min} \boldsymbol{\xi}^\intercal \mathbf{A} \boldsymbol{\xi} = \boldsymbol{\xi}^\intercal \mathbf{AS} \boldsymbol{\xi}$ because $\boldsymbol{\xi}^\intercal \mathbf{A} \boldsymbol{\xi} = 0$ by the skew-symmetry of $\mathbf{A}$. It therefore suffices to show that the inequality holds when $\sigma_{\min} = 0$ and $\kappa = \sigma_{\max}$.

Since $\mathbf{S}$ is positive semidefinite, there exists an upper-triangular square-root matrix $T$ such that $\mathbf{T}^\intercal \mathbf{T} = \mathbf{S}$ and so $\boldsymbol{\xi}^\intercal \mathbf{S} \boldsymbol{\xi} = \|\mathbf{T} \boldsymbol{\xi}\|^2$. Further,

$$|\boldsymbol{\xi}^\intercal \mathbf{AS} \boldsymbol{\xi}| \leq \|\mathbf{A}^\intercal \boldsymbol{\xi}\| \cdot \|\mathbf{T}^\intercal \mathbf{T} \boldsymbol{\xi}\| \leq \sqrt{\sigma_{\max}} \cdot \|\mathbf{A}^\intercal \boldsymbol{\xi}\| \cdot \|\mathbf{T} \boldsymbol{\xi}\|.$$

since $\|\mathbf{T}\|_2 = \sqrt{\sigma_{\max}}$. Putting the observations together obtains

$$\|\mathbf{T} \boldsymbol{\xi}\|^2 + \lambda (\|\mathbf{A} \boldsymbol{\xi}\|^2 - \langle \mathbf{A} \boldsymbol{\xi}, \mathbf{S} \boldsymbol{\xi} \rangle)$$
$$\geq \|\mathbf{T} \boldsymbol{\xi}\|^2 + \lambda (\|\mathbf{A} \boldsymbol{\xi}\|^2 - \|\mathbf{A} \boldsymbol{\xi}\| \|\mathbf{S} \boldsymbol{\xi}\|)$$
$$\geq \|\mathbf{T} \boldsymbol{\xi}\|^2 + \lambda \|\mathbf{A} \boldsymbol{\xi}\| (\|\mathbf{A} \boldsymbol{\xi}\| - \|\mathbf{S} \boldsymbol{\xi}\|)$$
$$\geq \|\mathbf{T} \boldsymbol{\xi}\|^2 + \lambda \|\mathbf{A} \boldsymbol{\xi}\| (\|\mathbf{A} \boldsymbol{\xi}\| - \sqrt{\sigma_{max}} \|\mathbf{T} \boldsymbol{\xi}\|)$$

Set $\alpha = \sqrt{\lambda}$ and $\eta = \sqrt{\sigma_{max}}$. We can continue the above computation

$$= \|\mathbf{T} \boldsymbol{\xi}\|^2 + \alpha^2 \|\mathbf{A} \boldsymbol{\xi}\| (\|\mathbf{A} \boldsymbol{\xi}\| - \eta \|\mathbf{T} \boldsymbol{\xi}\|)$$
$$= \|\mathbf{T} \boldsymbol{\xi}\|^2 + \alpha^2 \|\mathbf{A} \boldsymbol{\xi}\|^2 - \alpha^2 \|\mathbf{A} \boldsymbol{\xi}\| \eta \|\mathbf{T} \boldsymbol{\xi}\|$$
$$= (\|\mathbf{T} \boldsymbol{\xi}\| - \alpha \|\mathbf{A} \boldsymbol{\xi}\|)^2 + 2\alpha \|\mathbf{A} \boldsymbol{\xi}\| \|\mathbf{T} \boldsymbol{\xi}\| - \alpha^2 \eta \|A \boldsymbol{\xi}\| \|\mathbf{T} \boldsymbol{\xi}\|$$
$$= (\|\mathbf{T} \boldsymbol{\xi}\| - \alpha \|\mathbf{A} \boldsymbol{\xi}\|)^2 + \|\mathbf{A} \boldsymbol{\xi}\| \|\mathbf{T} \boldsymbol{\xi}\| (2\alpha - \alpha^2 \eta)$$

Finally, $2\alpha - \alpha^2 \eta > 0$ for any $\alpha$ in the range $(0, \frac{2}{\eta})$, which is to say, for any $0 < \lambda < \frac{4}{\sigma_{max}}$. The kernel of $\mathbf{S}$ and the kernel of $\mathbf{T}$ coincide. If $\boldsymbol{\xi}$ is in the kernel of $\mathbf{A}$, resp. $\mathbf{T}$, it cannot be in the kernel of $\mathbf{T}$, resp. $\mathbf{A}$ and the term $(\|\mathbf{T} \boldsymbol{\xi}\| - \alpha \|\mathbf{A} \boldsymbol{\xi}\|)^2$ is positive. Otherwise, the term $\|\mathbf{A} \boldsymbol{\xi}\| \|\mathbf{T} \boldsymbol{\xi}\|$ is positive. $\quad\square$

**Proof of proposition 6.**

**Lemma 10.** *When $\boldsymbol{\xi}_\lambda$ is the symplectic gradient adjustment,*

$$\text{sign} \Big( \text{align}(\boldsymbol{\xi}_\lambda, \nabla \mathcal{H}) \Big) = \text{sign} \Big( \langle \boldsymbol{\xi}, \nabla \mathcal{H} \rangle \cdot \langle \mathbf{A}^\intercal \boldsymbol{\xi}, \nabla \mathcal{H} \rangle \Big).$$

*Proof.* Observe that

$$\cos^2 \theta_\lambda = \left( \frac{\langle \boldsymbol{\xi}_\lambda, \nabla \mathcal{H} \rangle}{\|\boldsymbol{\xi}_\lambda\| \cdot \|\nabla \mathcal{H}\|} \right)^2$$
$$= \frac{\langle \boldsymbol{\xi}, \nabla \mathcal{H} \rangle + 2\lambda \langle \boldsymbol{\xi}, \nabla \mathcal{H} \rangle \langle \mathbf{A}^\intercal \boldsymbol{\xi}, \nabla \mathcal{H} \rangle + O(\lambda^2)}{(\|\boldsymbol{\xi}\|^2 + O(\lambda^2)) \cdot \|\nabla \mathcal{H}\|^2}$$

where the denominator has no linear term in $\lambda$ because $\boldsymbol{\xi} \perp \mathbf{A}^\intercal \boldsymbol{\xi}$. It follows that the sign of the infinitesimal alignment is

$$\text{sign} \left\{ \frac{d}{d\lambda} \cos^2 \theta_\lambda \right\} = \text{sign} \left\{ \langle \boldsymbol{\xi}, \nabla \mathcal{H} \rangle \langle \mathbf{A}^\intercal \boldsymbol{\xi}, \nabla \mathcal{H} \rangle \right\}$$

as required. $\quad\square$

The proposition follows easily.

*Proof.* If we are in a neighborhood of a stable fixed point then $\langle \boldsymbol{\xi}, \nabla \mathcal{H} \rangle \geq 0$. It follows by lemma 10 that $\text{sign} \Big( \text{align}(\boldsymbol{\xi}_\lambda), \nabla \mathcal{H} \Big) = \text{sign} \Big( \langle \mathbf{A}^\intercal \boldsymbol{\xi}, \nabla \mathcal{H} \rangle \Big)$ and so choosing $\text{sign}(\lambda) = \text{sign} \Big( \langle \mathbf{A}^\intercal \boldsymbol{\xi}, \nabla \mathcal{H} \rangle \Big)$ leads to the angle between $\boldsymbol{\xi}_\lambda$ and $\nabla \mathcal{H}$ being smaller than the angle between $\boldsymbol{\xi}$ and $\nabla \mathcal{H}$, satisfying desideratum $D4$.

The proof for the unstable case is similar. $\quad\square$

**Proof of theorem 7.** The following lemma provides some intuition for theorem 7. The idea is that, the smaller the cosine between the 'correct direction' $\mathbf{w}$ and the 'update direction' $\boldsymbol{\xi}$, the smaller the learning rate needs to be for the update to stay in a unit ball, see figure 6.

**Lemma 11** (alignment lemma)**.** *If $\mathbf{w}$ and $\boldsymbol{\xi}$ are unit vectors with $0 < \mathbf{w}^\intercal \boldsymbol{\xi}$ then $\|\mathbf{w} - \eta \cdot \boldsymbol{\xi}\| \leq 1$ for $0 \leq \eta \leq 2\mathbf{w}^\intercal \boldsymbol{\xi}$.*

*Proof.* Check $\|\mathbf{w} - \eta \cdot \boldsymbol{\xi}\|^2 = 1 + \eta^2 - 2\eta \cdot \mathbf{w} \cdot \boldsymbol{\xi} \leq 1$ iff $\eta^2 \leq 2\eta \cdot \mathbf{w}^\intercal \boldsymbol{\xi}$. The result follows. $\quad\square$

The next lemma is a standard technical result (e.g. Nesterov, 2004).

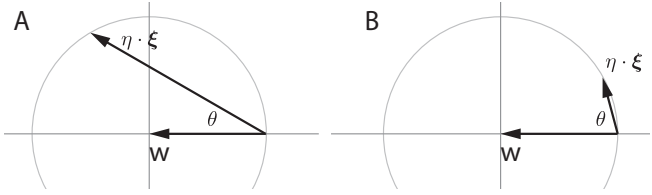

Figure 6. *Alignment and learning rates.* The larger $\cos\theta$, the larger the learning rate $\eta$ that can be applied to unit vector $\boldsymbol{\xi}$ without $\mathbf{w} + \eta \cdot \boldsymbol{\xi}$ leaving the unit circle.

**Lemma 12.** *Let $f : \mathbb{R}^d \to \mathbb{R}$ be a convex Lipschitz smooth function satisfying $\|\nabla f(\mathbf{y}) - \nabla f(\mathbf{x})\| \leq L \cdot \|\mathbf{y} - \mathbf{x}\|$ for all $\mathbf{x}, \mathbf{y} \in \mathbb{R}^d$. Then*

$$|f(\mathbf{y}) - f(\mathbf{x}) - \langle \nabla f(\mathbf{x}), \mathbf{y} - \mathbf{x} \rangle| \leq \frac{L}{2} \cdot \|\mathbf{y} - \mathbf{x}\|^2$$

*for all $\mathbf{x}, \mathbf{y} \in \mathbb{R}^d$.*

Finally the proof of the theorem is adapted from a standard result in Nesterov (2004), where there is no cosine.

*Proof.* By the above lemma,

$$\begin{aligned}
f(\mathbf{y}) &\leq f(\mathbf{x}) + \langle \nabla f(\mathbf{x}), \mathbf{y} - \mathbf{x} \rangle + \frac{L}{2}\|\mathbf{y} - \mathbf{x}\|^2 \\
&= f(\mathbf{x}) - \eta \cdot \langle \nabla f, \boldsymbol{\xi} \rangle + \eta^2 \frac{L}{2} \cdot \|\boldsymbol{\xi}\|^2 \\
&= f(\mathbf{x}) - \eta \cdot \langle \nabla f, \boldsymbol{\xi} \rangle + \eta^2 \frac{L}{2} \cdot \|\nabla f\|^2 \\
&= f(\mathbf{x}) - \eta(\alpha - \frac{\eta}{2}L) \cdot \|\nabla f\|^2
\end{aligned}$$

where $\alpha := \cos\theta$. Solve

$$\min_{\eta} \Delta(\eta) = \min_{\eta} \left\{ -\eta(\alpha - \frac{\eta}{2}L) \right\}$$

to obtain $\eta^* = \frac{\alpha}{L}$ and $\Delta(\eta^*) = -\frac{\alpha^2}{2}L$ as required. $\qquad\square$

## C. TensorFlow code to compute $\mathbf{A}^{\mathsf{T}}\boldsymbol{\xi}$:

The code requires a list of $n$ losses, `Ls`, and a list of variables for the $n$ players, `xs`. The function `fwd_gradients` is in the module `tf.contrib.kfac.utils`.

```
def jac_vec(ys,xs,vs):
  return fwd_gradients(ys,xs,
          grad_xs = vs, stop_gradients = xs)

def jac_tran_vec(ys,xs,vs):
  dydxs = tf.gradients(ys,xs,grad_ys = vs,
                  stop_gradients = xs)
  return [tf.zeros_like(x) if dydx is None
```

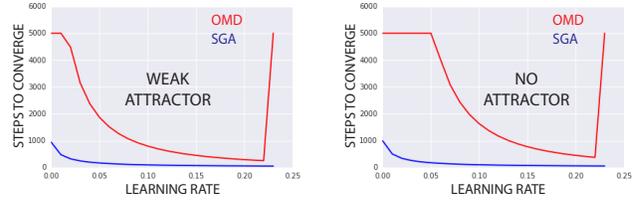

Figure 7. Time to convergence of OMD and SGA on two 4-player games. Times are cutoff after 5000 iterations. **Left panel:** Weakly positive definite $\mathbf{S}$ with $\epsilon = \frac{1}{100}$. **Right panel:** Symmetric component is identically zero.

```
        else dydx
        for (x,dydx) in zip(xs,dydxs)]

def get_sym_adj(Ls,xs):
  xi = [tf.gradients(ℓ,x)[0]
      for (ℓ,x) in zip(Ls,xs)]
  H_xi = jac_vec(xi,xs,xi)
  Ht_xi = jac_tran_vec(xi,xs,xi)
  At_xi = [ht-h/2 for (h,ht) in zip(H_xi,Ht_xi)]
  return At_xi
```

## D. Further experiments

**OMD and SGA on a four-player game.** Figure 7 shows time to convergence (same criterion as section 4.2 for optimistic mirror descent and SGA. The games are constructed with four players, each of which controls one parameter. The losses are

$$\begin{aligned}
\ell_1(w,x,y,z) &= \frac{\epsilon}{2}w^2 + wx + wy + wz \\
\ell_2(w,x,y,z) &= -wx + \frac{\epsilon}{2}x^2 + xy + xz \\
\ell_3(w,x,y,z) &= -wy - xy + \frac{\epsilon}{2}y^2 + yz \\
\ell_4(w,x,y,z) &= -wz - xz - yz + \frac{\epsilon}{2}z^2,
\end{aligned}$$

where $\epsilon = \frac{1}{100}$ in the left panel and $\epsilon = 0$ in the right panel. The antisymmetric component of the game Hessian is

$$\mathbf{A} = \begin{pmatrix} 0 & 1 & 1 & 1 \\ -1 & 0 & 1 & 1 \\ -1 & -1 & 0 & 1 \\ -1 & -1 & -1 & 0 \end{pmatrix}$$

and the symmetric component is

$$\mathbf{S} = \epsilon \cdot \begin{pmatrix} 1 & 0 & 0 & 0 \\ 0 & 1 & 0 & 0 \\ 0 & 0 & 1 & 0 \\ 0 & 0 & 0 & 1 \end{pmatrix}.$$

OMD converges considerably slower than SGA across the full range of learning rates. It also diverges for learning rates > 0.22. SGA converges more quickly and robustly.

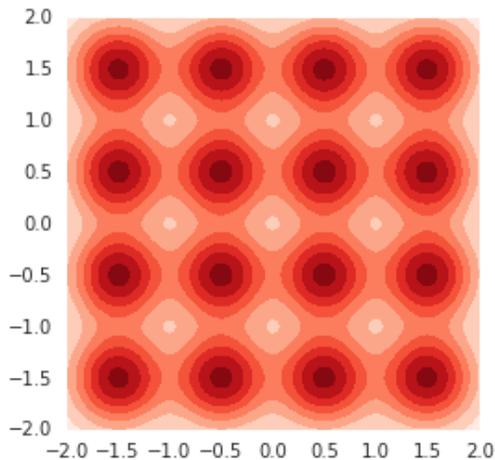

*Figure 8.* Ground truth for GAN experiments. A mixture of 16 Gaussians.

**Ground truth for GAN experiments.** Figure 8 shows the probability distribution that is sampled to train the generator and discriminator in the GAN example.

**More details on GAN experiments.** Figure 9 shows the performance of consensus optimization without and with alignment. Introducing alignment slightly improves speed of convergence (second column) and final result (fourth column), although intermediate results in third column are ambiguous.

Grid search was over learning rates {1e-5, 2e-5,5e-5, 8e-5, 1e-4, 2e-4, 5e-4} and then a more refined linear search over [8e-5, 2e-4].

# E. Helmholtz, Hamilton, Hodge, and Harmonic games

This section explains the mathematical connections with the Helmholtz decomposition, symplectic geometry and the Hodge decomposition. The discussion is *not* necessary to understand the main text. It is also not self-contained. The details can be found in textbooks covering differential and symplectic geometry.

## E.1. The Helmholtz decomposition

The classical Helmholtz decomposition states that any vector field $\boldsymbol{\xi}$ in 3-dimensions is the sum of curl-free (gradient) and divergence-free (infinitesimal rotation) components:

$$\boldsymbol{\xi} = \underbrace{\nabla\phi}_{\text{gradient component}} + \underbrace{\text{curl}(\mathbf{B})}_{\text{rotational component}} \qquad \left[\text{curl}(\bullet) := \nabla\times(\bullet)\right]$$

We explain the link between curl and the antisymmetric component of the game Hessian. Recall that gradients of functions are actually differential 1-forms, not vector fields.

Differential 1-forms and vector fields on a manifold are canonically isomorphic once a Riemannian metric has been chosen. In our case, we are implicitly using the Euclidean metric. The antisymmetric matrix $\mathbf{A}$ is the differential 2-form obtained by applying the exterior derivative $d$ to the 1-form $\boldsymbol{\xi}$.

In 3-dimensions, the Hodge star operator is an isormorphism from differential 2-forms to vector fields, and the curl can be reformulated as $\text{curl}(\bullet) = *d(\bullet)$. In claiming $\mathbf{A}$ is analogous to curl, we are simply dropping the Hodge-star operator.

Finally, recall that the Lie algebra of infinitesimal rotations in $d$-dimensions is given by antisymmetric matrices. When $d = 3$, the Lie algebra can be represented as vectors (three numbers specify a $3 \times 3$ antisymmetric matrix) with the $\times$-product as Lie bracket. In general, the antisymmetric matrix $\mathbf{A}$ captures the infinitesimal tendency of $\boldsymbol{\xi}$ to rotate at each point in the parameter space.

## E.2. Hamiltonian mechanics

A symplectic form $\omega$ is a closed nondegenerate differential 2-form. Given a manifold with a symplectic form, a vector field $\boldsymbol{\xi}$ is **Hamiltonian vector field** if there exists a function $\mathcal{H} : M \to \mathbb{R}$ satisfying

$$\omega(\boldsymbol{\xi}, \bullet) = d\mathcal{H}(\bullet) = \langle\nabla\mathcal{H}, \bullet\rangle. \qquad (3)$$

The function is then referred to as the Hamiltonian function of the vector field. In our case, the antisymmetric matrix $\mathbf{A}$ is a closed 2-form because $\mathbf{A} = d\boldsymbol{\xi}$ and $d \circ d = 0$. It may however be degenerate. It is therefore a presymplectic form (Bottacin, 2005).

Setting $\omega = \mathbf{A}$, equation (3) can be rewritten in our notation as

$$\underbrace{\omega(\boldsymbol{\xi}, \bullet)}_{\mathbf{A}^\top\boldsymbol{\xi}} = \underbrace{d\mathcal{H}(\bullet)}_{\nabla\mathcal{H}},$$

justifying the terminology 'Hamiltonian'.

## E.3. The Hodge decomposition

The exterior derivative $d_k : \Omega^k(M) \to \Omega^{k+1}(M)$ is a linear operator that takes differential $k$-forms on a manifold $M$, $\Omega^k(M)$, to differential $k + 1$-forms, $\Omega^{k+1}(M)$. In the case $k = 0$, the exterior derivative is the gradient, which takes 0-forms (that is, functions) to 1-forms. Given a Riemannian metric, the adjoint of the exterior derivative $\delta$ goes in the opposite direction. Hodge's theorem states that $k$-forms on a compact manifold decompose into a direct sum over three types:

$$\Omega^k(M) = d\Omega^{k-1}(M) \oplus \text{Harmonic}^k(M) \oplus \delta\Omega^{k+1}(M).$$

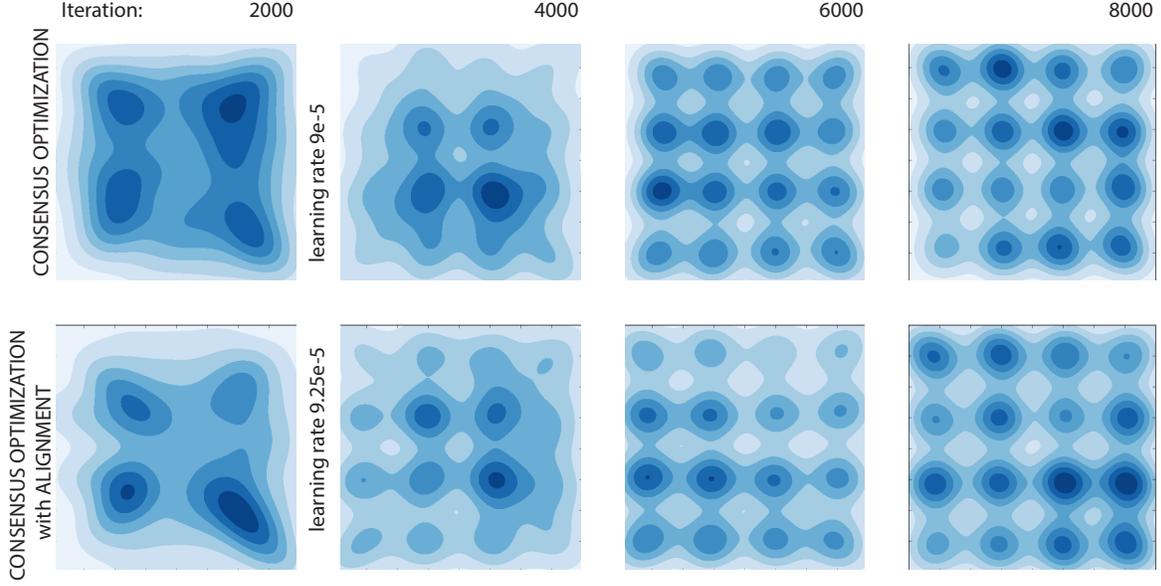

*Figure 9.* **Top:** Consensus optimization on GAN example. **Bottom:** Consensus optimization with alignment.

Setting $k = 1$, we recover a decomposition that closely resembles the generalized Helmholtz decomposition:

$$\underbrace{\Omega^1(M)}_{\text{1-forms}} = \underbrace{d\Omega^0(M)}_{\text{gradients of functions}} \oplus \text{Harm}^k(M) \oplus \underbrace{\delta\Omega^2(M)}_{\text{antisymmetric component}}$$

The harmonic component is isomorphic to the de Rham cohomology of the manifold – which is zero when $k = 1$ and $M = \mathbb{R}^n$.

Unfortunately, the Hodge decomposition does not straightforwardly apply to the case when $M = \mathbb{R}^n$, since $\mathbb{R}^n$ is not compact. It is thus unclear how to relate the generalized Helmholtz decomposition to the Hodge decomposition.

### E.4. Harmonic and potential games

Candogan et al. (2011) derive a Hodge decomposition for games that is closely related in spirit to our generalized Helmholtz decomposition – although the details are quite different. Candogan et al. (2011) work with classical games (probability distributions on finite strategy sets). Their losses are multilinear, which is easier than our setting, but they have constrained solution sets, which is harder in many ways. Their approach is based on combinatorial Hodge theory rather than differential and symplectic geometry. Finding a best-of-both-worlds approach that encompasses both settings is an open problem.

## F. Type consistency

The next two sections carefully work through the units in classical mechanics and two-player games respectively. The third section briefly describes a use-case for type consistency.

### F.1. Units in classical mechanics

Consider the well-known Hamiltonian

$$\mathcal{H}(p, q) = \frac{1}{2}\left(\kappa \cdot q^2 + \frac{1}{\mu} \cdot p^2\right)$$

where $q$ is position, $p = \mu \cdot \dot{q}$ is momentum, $\mu$ is mass, $\kappa$ is surface tension and $\mathcal{H}$ measures energy. The units (denoted by $\tau$) are

$$\tau(q) = m \qquad \tau(p) = \frac{kg \cdot m}{s}$$

$$\tau(\kappa) = \frac{kg}{s^2} \qquad \tau(\mu) = kg$$

where $m$ is meters, $kg$ is kilograms and $s$ is seconds. Energy is measured in joules, and indeed it is easy to check that $\tau(\mathcal{H}) = \frac{kg \cdot m^2}{s^2}$.

Note that the units for differentation by $x$ are $\tau(\frac{\partial}{\partial x}) = \frac{1}{\tau(x)}$. For example, differentiating by time has units $\frac{1}{s}$. Hamilton's equations state that $\dot{q} = \frac{\partial H}{\partial p} = \frac{1}{\mu} \cdot p$ and $\dot{p} = -\frac{\partial \mathcal{H}}{\partial q} = -\kappa \cdot q$ where

$$\tau(\dot{q}) = \frac{m}{s} \qquad \tau(\dot{p}) = \frac{kg \cdot m}{s^2}$$

$$\tau\left(\frac{\partial}{\partial q}\right) = \frac{1}{m} \qquad \tau\left(\frac{\partial}{\partial p}\right) = \frac{s}{kg \cdot m}$$

The resulting flow describing the dynamics of the system is

$$\boldsymbol{\xi} = \dot{q} \cdot \frac{\partial}{\partial q} + \dot{p} \cdot \frac{\partial}{\partial p} = \frac{1}{\mu} p \cdot \frac{\partial}{\partial q} - \kappa q \cdot \frac{\partial}{\partial p}$$

with units $\tau(\boldsymbol{\xi}) = \frac{1}{s}$. Hamilton's equations can be reformulated more abstractly via symplectic geometry. Introduce

the symplectic form

$$\omega = dq \wedge dp \quad \text{with units} \quad \tau(\omega) = \frac{kg \cdot m^2}{s}.$$

Observe that contracting the flow with the Hamiltonian obtains

$$\iota_{\boldsymbol{\xi}} \omega = \omega(\boldsymbol{\xi}, \bullet) = dH = \frac{\partial \mathcal{H}}{\partial q} \cdot dq + \frac{\partial \mathcal{H}}{\partial p} \cdot dp$$

with units $\tau(d\mathcal{H}) = \tau(\mathcal{H}) = \frac{kg \cdot m^2}{s^2}$.

**Losses in classical mechanics.** Although there is no notion of "loss" in classical mechanics, it is useful (for the next section) to keep pushing the formal analogy. Define the "losses"

$$\ell_1(q,p) = \frac{1}{\mu} \cdot qp \quad \text{and} \quad \ell_2(q,p) = -\kappa \cdot qp \quad (4)$$

with units $\tau(\ell_1) = \frac{m^2}{s}$ and $\tau(\ell_2) = \frac{kg^2 \cdot m^2}{s^3}$. The Hamiltonian dynamics can then be recovered game-theoretically by differentiating $\ell_1$ and $\ell_2$ with respect to $q$ and $p$ respectively. It is easy to check that

$$\boldsymbol{\xi} = \frac{\partial \mathcal{H}}{\partial p} \frac{\partial}{\partial q} - \frac{\partial \mathcal{H}}{\partial q} \frac{\partial}{\partial p} = \frac{\partial \ell_1}{\partial q} \frac{\partial}{\partial q} + \frac{\partial \ell_2}{\partial p} \frac{\partial}{\partial p}.$$

**The duality between vector fields and differential forms.** Finally recall that the symplectic form in games was not "pulled out of thin air" as $\omega = dq \wedge dp$, but rather derived as $\omega = d\boldsymbol{\xi}^\flat$, where $\boldsymbol{\xi}^\flat$ is the differential form corresponding to the vector field $\boldsymbol{\xi}$ under the musical isomorphism $\flat : TM \to T^*M$.

It is instructive to compute $\boldsymbol{\xi}^\flat$ in the case of a classical mechanical system and see what happens. Naively, we would guess that the musical isomorphism is $\left(\frac{\partial}{\partial q}\right)^\flat = dq$ and $\left(\frac{\partial}{\partial p}\right)^\flat = dp$. However, applying the naive musical isomorphism to $\boldsymbol{\xi}$ to get

$$\boldsymbol{\xi}^\flat = \frac{\partial \ell_1}{\partial q} \cdot dq + \frac{\partial \ell_2}{\partial p} \cdot dp$$

results in a *type violation* because

$$\tau\left(\frac{\partial \ell_1}{\partial q} \cdot dq\right) = \tau(\ell_1) = \frac{m^2}{s}$$

whereas

$$\tau\left(\frac{\partial \ell_2}{\partial p} \cdot dp\right) = \tau(\ell_2) = \frac{kg^2 \cdot m^2}{s^3}$$

and we cannot add objects with different types.

To correct the type inconsistency, define the musical isomorphism as

$$\left(\frac{\partial}{\partial q}\right)^\flat = \frac{\mu}{2} \cdot dq \quad \text{and} \quad \left(\frac{\partial}{\partial p}\right)^\flat = \frac{1}{2\kappa} \cdot dp$$

with inverse

$$(dq)^\sharp = \frac{2}{\mu} \cdot \frac{\partial}{\partial q} \quad \text{and} \quad (dp)^\sharp = 2\kappa \cdot \frac{\partial}{\partial p}.$$

The correction terms in the direction $\flat : TM \to T^*M$ invert the coupling terms $\kappa$ and $\frac{1}{\mu}$ that were originally introduced into the Hamiltonian for physical reasons. Applying the corrected musical isomorphism to $\boldsymbol{\xi}$ yields

$$\boldsymbol{\xi}^\flat = \frac{\mu}{2} \cdot \frac{\partial f}{\partial q} \cdot dq + \frac{1}{2\kappa} \cdot \frac{\partial g}{\partial p} \cdot dp = \frac{1}{2} \left( p \cdot dq - q \cdot dp \right).$$

The two terms of $\boldsymbol{\xi}^\flat$ then have coherent types

$$\tau\left(\frac{\partial \ell_1}{\partial q} \cdot \mu \cdot dq\right) = \frac{m}{s} \cdot kg \cdot m = \frac{kg \cdot m^2}{s}$$
$$\tau\left(\frac{\partial \ell_2}{\partial p} \cdot \frac{1}{\kappa} \cdot dp\right) = \frac{kg \cdot m}{s^2} \cdot \frac{s^2}{kg} \cdot \frac{kg \cdot m}{s} = \frac{kg \cdot m^2}{s}$$

as required. The associated two form is

$$\omega := d\boldsymbol{\xi}^\flat = -\left(\mu \cdot \frac{\partial^2 f}{\partial q \partial p} - \frac{1}{\kappa} \cdot \frac{\partial^2 g}{\partial q \partial p}\right) dq \wedge dp = -dq \wedge dp$$

which recovers the symplectic form (up to sign) with units $\tau(\omega) = \frac{kg \cdot m^2}{s}$ as required. Finally, observe that

$$\langle \boldsymbol{\xi}, \boldsymbol{\xi}^\flat \rangle = \frac{1}{2} \left\langle \frac{p}{\mu} \cdot \frac{\partial}{\partial q} - \kappa q \cdot \frac{\partial}{\partial p}, p \cdot dq - q \cdot dp \right\rangle$$
$$= \frac{1}{2} \left( \kappa \cdot q^2 + \frac{1}{\mu} \cdot p^2 \right) = \mathcal{H}(p,q)$$

recovering the Hamiltonian.

## F.2. Units in two-player games

Without loss of generality let $\mathbf{w} = (\mathbf{x}; \mathbf{y})$ where we refer to $\mathbf{x}$ as position and $\mathbf{y}$ as momentum so that $\tau(\mathbf{x}) = m$ and $\tau(\mathbf{y}) = \frac{kg \cdot m}{s}$. The aim of this section is to check type-consistency under these, rather arbitrarily assigned, units. Since we are considering a game, we do not require that $\mathbf{x}$ and $\mathbf{y}$ have the same dimension – even though this would necessarily be the case for a physical system. The goal is to verify that units can be consistently assigned to games.

Consider a quadratic two player game of the form

$$\ell_1(\mathbf{w}) = \frac{1}{2} \begin{pmatrix} \mathbf{x}^\mathsf{T} & \mathbf{y}^\mathsf{T} \end{pmatrix} \begin{pmatrix} \mathbf{A}_{11} & \mathbf{A}_{12} \\ \mathbf{A}_{21} & \mathbf{A}_{22} \end{pmatrix} \begin{pmatrix} \mathbf{x} \\ \mathbf{y} \end{pmatrix} + \begin{pmatrix} \mathbf{x}^\mathsf{T} & \mathbf{y}^\mathsf{T} \end{pmatrix} \begin{pmatrix} \mathbf{b}_1 \\ \mathbf{b}_2 \end{pmatrix}$$

and

$$\ell_2(\mathbf{w}) = \frac{1}{2} \begin{pmatrix} \mathbf{x}^\mathsf{T} & \mathbf{y}^\mathsf{T} \end{pmatrix} \begin{pmatrix} \mathbf{C}_{11} & \mathbf{C}_{12} \\ \mathbf{C}_{21} & \mathbf{C}_{22} \end{pmatrix} \begin{pmatrix} \mathbf{x} \\ \mathbf{y} \end{pmatrix} + \begin{pmatrix} \mathbf{x}^\mathsf{T} & \mathbf{y}^\mathsf{T} \end{pmatrix} \begin{pmatrix} \mathbf{d}_1 \\ \mathbf{d}_2 \end{pmatrix}$$

We restrict to quadratic games since our methods only involve first and second derivatives. We assume the matrices $\mathbf{A}$ and $\mathbf{C}$ are symmetric without loss of generality so that, for example, $\mathbf{A}_{12} = \mathbf{A}_{21}^{\mathsf{T}}$. Adding constant terms to $\ell_1$ and $\ell_2$ makes no difference to the analysis so they are omitted.

By (4), the units for $\ell_1$ and $\ell_2$ should be $\frac{m^2}{s}$ and $\frac{kg \cdot m^2}{s^3}$ respectively. We can therefore derive the correct units for each of the components of the quadratic losses as

$$\underbrace{\left(m \ \tfrac{kg \cdot m}{s}\right) \begin{pmatrix} \frac{1}{s} & \frac{1}{kg} \\ \frac{1}{kg} & \frac{s}{kg^2} \end{pmatrix} \begin{pmatrix} m \\ \frac{kg \cdot m}{s} \end{pmatrix}}_{\mathbf{w}^{\mathsf{T}} \mathbf{A} \mathbf{w}} + \underbrace{\left(m \ \tfrac{kg \cdot m}{s}\right) \begin{pmatrix} \frac{m}{s} \\ \frac{m}{kg} \end{pmatrix}}_{\mathbf{w}^{\mathsf{T}} \mathbf{b}}$$

for $\ell_1$ and

$$\underbrace{\left(m \ \tfrac{kg \cdot m}{s}\right) \begin{pmatrix} \frac{kg^2}{s^3} & \frac{kg}{s^2} \\ \frac{kg}{s^2} & \frac{1}{s} \end{pmatrix} \begin{pmatrix} m \\ \frac{kg \cdot m}{s} \end{pmatrix}}_{\mathbf{w}^{\mathsf{T}} \mathbf{C} \mathbf{w}} + \underbrace{\left(m \ \tfrac{kg \cdot m}{s}\right) \begin{pmatrix} \frac{kg^2 \cdot m}{s^3} \\ \frac{kg \cdot m}{s^2} \end{pmatrix}}_{\mathbf{w}^{\mathsf{T}} \mathbf{d}}$$

for $\ell_2$. It follows from a straightforward computation that the vector field $\boldsymbol{\xi} = \frac{\partial \ell_1}{\partial \mathbf{x}} \frac{\partial}{\partial \mathbf{x}} + \frac{\partial \ell_2}{\partial \mathbf{y}} \frac{\partial}{\partial \mathbf{y}}$ has type $\tau(\boldsymbol{\xi}) = \frac{1}{s}$ as required.

The presymplectic form $\omega = d\boldsymbol{\xi}^{\flat}$ makes use of the musical isomorphism $\flat : T^M \to T^* M$. As in section F.1, if we naively define $\left(\frac{\partial}{\partial \mathbf{x}}\right)^{\flat} = d\mathbf{x}$ and $\left(\frac{\partial}{\partial \mathbf{y}}\right)^{\flat} = d\mathbf{y}$ then

$$\boldsymbol{\xi}^{\flat} = \frac{\partial \ell_1}{\partial \mathbf{x}} \cdot d\mathbf{x} + \frac{\partial \ell_2}{\partial \mathbf{y}} \cdot d\mathbf{y}$$

which is type inconsistent because $\tau(\frac{\partial \ell_1}{\partial \mathbf{x}} \cdot d\mathbf{x}) = \frac{m^2}{s}$ and $\tau(\frac{\partial \ell_2}{\partial \mathbf{y}} \cdot d\mathbf{y}) = \frac{kg^2 \cdot m^2}{s^3}$.

**Type-consistency via SVD.** It is necessary, as in section F.1, to correct the naive musical isomorphism by taking into account the coupling constants for the mixed position-momentum terms. In the classical setup the coupling constants were the scalars $\frac{1}{\mu}$ and $\kappa$, whereas in a game they are the off-diagonal blocks $\mathbf{A}_{12}$ and $\mathbf{C}_{12}$.

Apply singular value decomposition to factorize

$$\mathbf{A}_{12} = \mathbf{U}_{\mathbf{A}}^{\mathsf{T}} \mathbf{D}_{\mathbf{A}} \mathbf{V}_{\mathbf{A}} \quad \text{and} \quad \mathbf{C}_{12} = \mathbf{U}_{\mathbf{C}}^{\mathsf{T}} \mathbf{D}_{\mathbf{C}} \mathbf{V}_{\mathbf{C}}$$

where the entries of the diagonal matrices have types $\tau(\mathbf{D}_{\mathbf{A}}) = \frac{1}{kg}$ and $\tau(\mathbf{D}_{\mathbf{C}}) = \frac{kg}{s^2}$, and *the types of the orthogonal matrices* $\mathbf{U}$ *and* $\mathbf{V}$ *are pure scalars*. The diagonal matrices $\mathbf{D}_{\mathbf{A}}$ and $\mathbf{D}_{\mathbf{C}}$ have the same types as $\frac{1}{\mu}$ and $\kappa$ in the classical system since they play the same coupling role.

Extending the procedure adopted in the section F.1, fix the type-inconsistency by defining the musical isomorphisms as

$$\left(\frac{\partial}{\partial \mathbf{x}}\right)^{\flat} = \mathbf{U}_{\mathbf{A}}^{\mathsf{T}} \mathbf{D}_{\mathbf{A}}^{-1} \mathbf{U}_{\mathbf{A}} \cdot d\mathbf{x}$$

and

$$\left(\frac{\partial}{\partial \mathbf{y}}\right)^{\flat} = \mathbf{V}_{\mathbf{C}}^{\mathsf{T}} \mathbf{D}_{\mathbf{C}}^{-1} \mathbf{V}_{\mathbf{C}} \cdot d\mathbf{y}.$$

Alternatively, the isomorphisms can be computed by noting that $\mathbf{U}_{\mathbf{A}}^{\mathsf{T}} \mathbf{D}_{\mathbf{A}}^{-1} \mathbf{U}_{\mathbf{A}} = (\sqrt{\mathbf{A}_{12} \mathbf{A}_{21}})^{-1}$ and $\mathbf{V}_{\mathbf{C}}^{\mathsf{T}} \mathbf{D}_{\mathbf{C}}^{-1} \mathbf{V}_{\mathbf{C}} = (\sqrt{\mathbf{C}_{21} \mathbf{C}_{12}})^{-1}$.

The dual isomorphism $\sharp : T^* M \to TM$ is then

$$(d\mathbf{x})^{\sharp} = \mathbf{U}_{\mathbf{A}}^{\mathsf{T}} \mathbf{D}_{\mathbf{A}} \mathbf{U}_{\mathbf{A}} \cdot \frac{\partial}{\partial \mathbf{x}} \quad \text{and} \quad (d\mathbf{y})^{\sharp} = \mathbf{V}_{\mathbf{C}}^{\mathsf{T}} \mathbf{D}_{\mathbf{C}} \mathbf{V}_{\mathbf{C}} \cdot \frac{\partial}{\partial \mathbf{y}}$$

If

$$\boldsymbol{\xi} = \begin{pmatrix} \mathbf{A}_{12} \mathbf{y} + \mathbf{b}_1 \\ \mathbf{C}_{21} \mathbf{x} + \mathbf{d}_2 \end{pmatrix}$$

then it follows that

$$\boldsymbol{\xi}^{\flat} = \begin{pmatrix} \mathbf{U}_{\mathbf{A}}^{\mathsf{T}} \mathbf{D}_{\mathbf{A}}^{-1} \mathbf{U}_{\mathbf{A}} \mathbf{b}_1 + \mathbf{U}_{\mathbf{A}}^{\mathsf{T}} \mathbf{U}_{\mathbf{A}} \mathbf{y} \\ \mathbf{V}_{\mathbf{C}}^{\mathsf{T}} \mathbf{D}_{\mathbf{C}}^{-1} \mathbf{V}_{\mathbf{C}} \mathbf{d}_2 + \mathbf{V}_{\mathbf{C}}^{\mathsf{T}} \mathbf{V}_{\mathbf{C}} \mathbf{x} \end{pmatrix}$$

with associated closed two form

$$\omega_{\tau} = d\boldsymbol{\xi}^{\flat} = -\left(\mathbf{U}_{\mathbf{A}}^{\mathsf{T}} \mathbf{V}_{\mathbf{A}} - \mathbf{U}_{\mathbf{C}}^{\mathsf{T}} \mathbf{V}_{\mathbf{C}}\right) d\mathbf{x} \wedge d\mathbf{y}.$$

where the notation $\omega_{\tau}$ emphasizes that the two-form is type-consistent.

### F.3. What does type-consistency buy?

**Example 8.** *Consider the loss functions*

$$f(x, y) = xy \quad \text{and} \quad g(x, y) = 2xy,$$

*with* $\boldsymbol{\xi} = (y, 2x)$. *There is no function* $\phi : \mathbb{R}^2 \to \mathbb{R}$ *such that* $\nabla \phi = \boldsymbol{\xi}$. *However, there is a family of functions* $\phi_{\alpha}(x, y) = \alpha \cdot xy$ *which satisfies*

$$\langle \boldsymbol{\xi}, \nabla \phi_{\alpha} \rangle = \alpha \cdot (x^2 + 2y^2) \geq 0 \quad \text{for all } \alpha > 0.$$

Although $\boldsymbol{\xi}$ is not a potential field, there is a family of functions on which $\boldsymbol{\xi}$ performs gradient descent – albeit with coordinate-wise learning rates that may not be optimal. The vector field $\boldsymbol{\xi}$ arguably does not require adjustment. This kind of situation often arises when the learning rates of different parameters are set adaptively during training of neural nets, by rescaling them by positive numbers.

The vanilla and type-consistent 1-forms corresponding to $\boldsymbol{\xi}$ are, respectively,

$$\boldsymbol{\xi}^{\flat} = y \cdot dx + 2x \cdot dy \quad \text{and} \quad \boldsymbol{\xi}_{\tau}^{\flat} = y \cdot dx + x \cdot dy$$

with

$$\omega = d\boldsymbol{\xi}_{\text{non}}^{\flat} = dx \wedge dy \quad \text{and} \quad \omega_{\tau} = d\boldsymbol{\xi}_{\tau}^{\flat} = 0.$$

It follows that the *type-consistent* symplectic gradient adjustment is zero. Type-consistency 'detects' that no gradient adjustment is needed in example 8.

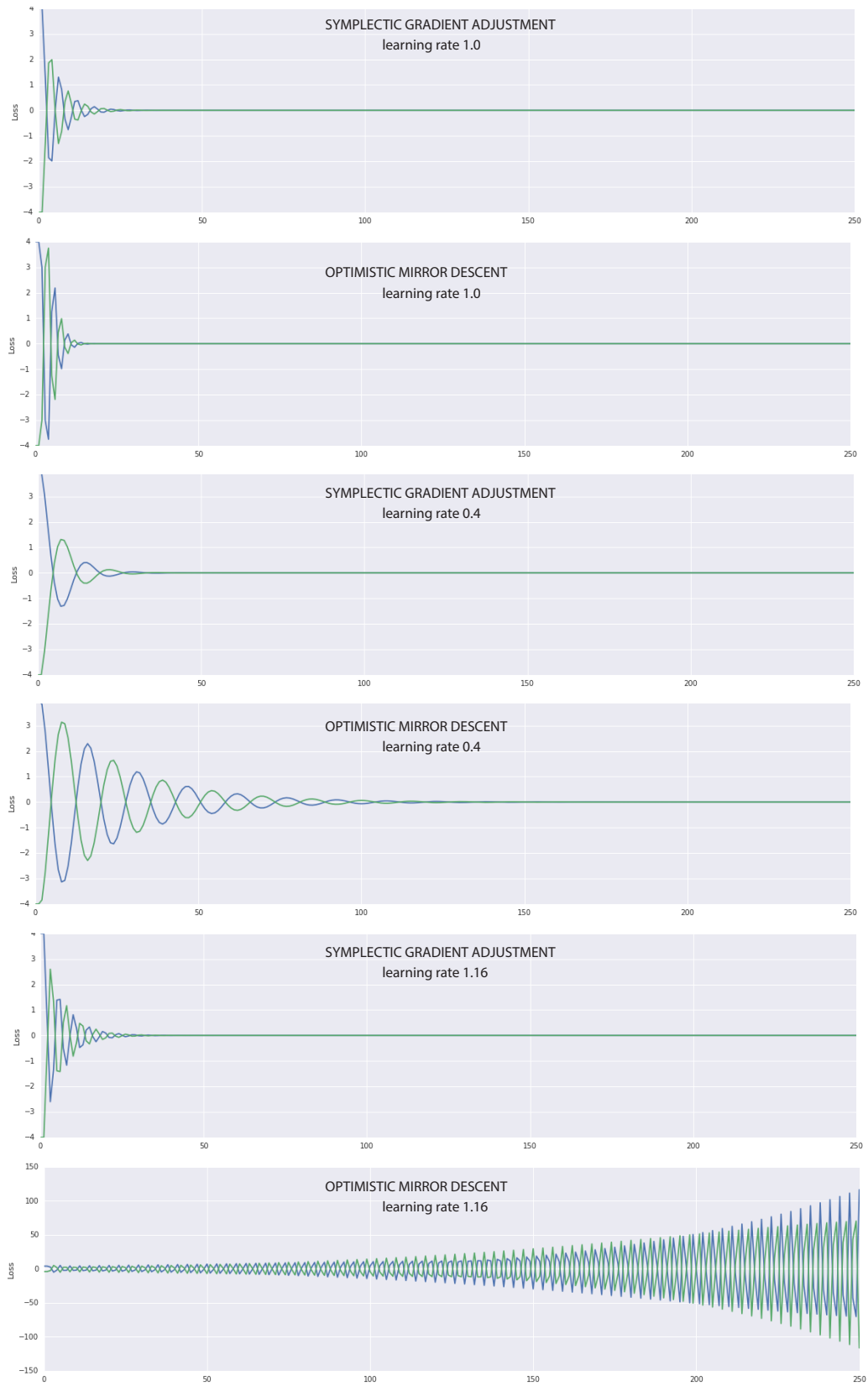

*Figure 10.* Individual runs on zero-sum bimatrix game in section 4.2.